\documentclass[runningheads]{llncs}
\usepackage[T1]{fontenc}

\usepackage{amsmath}
\usepackage{amssymb}

\usepackage{amsthm}
\usepackage{amsthm}    % Uncomment if you define theorems, lemmas, etc.
\usepackage{latexsym}
\usepackage{bm}

\usepackage{graphicx}
\usepackage{caption}
\usepackage{subcaption}
\usepackage{booktabs}
\usepackage{multirow}
\usepackage{mwe}        % Example images (optional)

\usepackage{algorithm}
\usepackage{algpseudocode}

\usepackage{pifont}     % For symbols like checkmarks
\usepackage[misc]{ifsym}
\newcommand{\corr}{(\Letter)}  % Corresponding author symbol
\usepackage{enumitem}   % Customizable list environments
\usepackage{color}
\usepackage{balance}    % For balancing columns on the final page
\usepackage{pdflscape}  % For rotating specific pages
\usepackage{hyperref}   % Clickable references and URLs

\usepackage{geometry}
\usepackage{algorithm}

\usepackage{algpseudocode}
\usepackage{geometry}
\usepackage{graphicx}
\usepackage{subcaption}

\begin{document}

\title{UCB-driven Utility Function Search for Multi-objective Reinforcement Learning}

\titlerunning{UCB-driven MOPPO}
% If the full title of your paper is short enough to also fit in the running head, you can omit the abbreviated paper title here. You can check as follows: if you comment out the \titlerunning line, something will appear in the header of all odd-numbered pages of your PDF from page 3 onward. This something is either the full title (in which case all is well), or the error message "Title Suppressed Due to Excessive Length". If this error message appears, you're going to want to provide an abbreviated title within the \titlerunning command, because if you won't do it, Springer will do it for you.

%N.B.: Author information (both in the \author{} and \authorrunning{} command) should only be present in the Camera-Ready Version of your paper. The version that you initially submit for review, ought to be double-blind. So, when initially submitting your paper, use:
% \author{Author information scrubbed for double-blind reviewing}

\author{Yucheng Shi \inst{1}\and
David Lynch \inst{2} \and
Alexandros Agapitos \inst{2} \corr }
% % You may leave out the orcidID information, if you want to.
% % Use \corr to indicate the corresponding author. Note the spacing around the \corr command. Only one author can be the corresponding author.

% %N.B.: comment out the \authorrunning{} command for the double-blind version of your paper submitted for review. Later, if your paper is accepted, use the command for the Camera-Ready Version.

\authorrunning{Yucheng et al.}
% % If there is one author, write 'A.L. Benjamin'.
% % If there are two authors, write 'A.L. Benjamin and C.C. Broadus Jr.'
% % If there are more than two authors, '[...] et al.' is used.

\institute{Trinity College Dublin, Ireland \email{shiy2@tcd.ie}
\and
Huawei Ireland Research Centre, Ireland \email{\{david.lynch1, alexandros.agapitos\}@huawei.com}
}

\maketitle              % typeset the header of the contribution

\begin{abstract}
In Multi-objective Reinforcement Learning (MORL) agents are tasked with optimising decision-making behaviours that trade-off between multiple, possibly conflicting, objectives. MORL based on decomposition is a family of solution methods that employ a number of utility functions to decompose the multi-objective problem into individual single-objective problems solved simultaneously in order to approximate a Pareto front of policies. We focus on the case of linear utility functions parametrised by weight vectors \textbf{w}. We introduce a method based on Upper Confidence Bound to efficiently search for the most promising weight vectors during different stages of the learning process, with the aim of maximising the hypervolume of the resulting Pareto front. The proposed method demonstrates consistency and strong performance across various MORL baselines on Mujoco benchmark problems. The code is released in: https://github.com/SYCAMORE-1/ucb-MOPPO

\end{abstract}

\section{Introduction}
\label{sec:Introduction}
In many real-world control and planning problems, multiple and often conflicting objectives arise. These objectives are interrelated, requiring trade-offs that significantly affect the overall quality of decision-making ~\cite{vamplew2011empirical,roijers2013survey}. Such learning objectives are usually represented as weighted reward signals, where conflicting rewards can lead to divergent optimisation directions \cite{zhang2020convex}. Consequently, the classic reinforcement learning (RL) methods are inadequate, as training individual policies to align with each preference weight vector across multiple rewards results in an impractical computational burden ~\cite{hayes2022practical,felten2024multi}.  Therefore, Multi-objective reinforcement learning (MORL) has become an increasingly recognised methodology for handling tasks with conflicting objectives, enabling the simultaneous optimisation of multiple criteria \cite{xu2020reinforcement,basaklar2022multiobjective,zhu2023survey}. In cases where the objectives conflict with each other,  the best trade-offs among the objectives are defined in terms of Pareto optimal. A policy $\pi$ is said to dominate $\pi'$ if $\big(\forall\,\,i : \textbf{V}{_i}^{\pi} \geq \textbf{V}{_i}^{\pi'}\big)$ $\land$ $\big(\exists\,\,i : \textbf{V}{_i}^{\pi} > \textbf{V}{_i}^{\pi'}\big)$, and it is Pareto optimal if there is no other policy that dominates it. The set of Pareto optimal policies is called the Pareto front (PF). In many real-world applications, an approximation to the PF is required by a decision maker in order to select a preferred policy~\cite{hayes2022practical}. A real-world use case is the multi-objective optimisation of wireless networks~\cite{gao2023coverage}. In this case, four primary objective classes, i.e.,throughput, coverage, energy efficiency and utilisation are considered. This is a very challenging problem that motivates further research to advance the state-of-the-art (SOTA) in MORL.

One prevalent category of MORL approaches is Multi-Objective Reinforcement Learning with Decomposition (MORL/D)~\cite{felten2024multi}. A Pareto-optimal solution to a multi-objective sequential decision-making problem can be interpreted as an optimal solution to a single-objective problem, defined with respect to a utility function \( u : \mathbb{R}^m \rightarrow \mathbb{R} \). This utility function aggregates \( m \) objectives into a scalar reward, mapping the multi-objective value of the policy into a scalar value, expressed as \( V_u^{\pi} = u (\textbf{V}^{\pi}) \)~\cite{Feinberg1995ConstrainedMD}.
Therefore, the approximation of the PF can be reformulated as a set of scalar reward sub-problems, each defined by a utility function. This approach unifies MORL methods~\cite{hayes2022practical} with decomposition-based multi-objective optimisation techniques~\cite{zhang2007moead}, collectively termed MORL/D.

MORL/D offers several advantages, including scalability to many objectives, flexibility in employing various scalarisation techniques, and the ability to parallelise sub-problem training, thereby reducing computational overhead~\cite{Felten2023ATF,felten2024multi}. SOTA MORL/D methods can be broadly categorised into single-policy and multi-policy approaches~\cite{felten2024multi}. Single-policy methods~\cite{abels2019dynamic,alegre2023sample,Lu2023MultiObjectiveRL,Yang2019AGA}  aim to learn a single policy conditioned on the parameters of the utility function, i.e., scalarisation vector \textbf{w}, where $\forall i: w_i \geq 0$, and $\sum_i w_i = 1.0$. However, these methods face two major challenges: (1) single-policy methods require large neural networks and extensive training time to accurately reconstruct the optimal PF; and (2) they may struggle to generalise effectively to unseen \textbf{w}~\cite{yang2019generalized}.  In contrast, multi-policy approaches maintain a separate policy for each \textbf{w}~\cite{Liu2021PredictionGM,mossalam2016multiobjective}. Parameter sharing improves the sample efficiency of single-policy methods and enhances generalisation to new objective preferences~\cite{abels2019dynamic,Yang2019AGA}. However, multi-policy methods often require maintaining various suboptimal policies to adequately cover the preference space, which can be highly memory-intensive ~\cite{Liu2021PredictionGM}.  Existing MORL/D approaches share common limitations. As the number of objectives increases or the granularity of the weight space becomes finer, the total preference space grows exponentially. This exponential growth renders prefixed scalarisation weights inefficient and degrades the quality of the approximated PF.

To address the limitations of MORL/D, particularly in the multi-policy paradigm, this work introduces an Upper Confidence Bound (UCB) acquisition function to identify promising scalarisation vectors $\textbf{w} \in \textbf{W}$ at different learning stages, guided by the Pareto front quality metric of \emph{hypervolume} (HV) ~\cite{Felten2023ATF}. Extending~\cite{abels2019dynamic}, we use a weight-conditioned Actor-Critic network $\pi_{\theta}(s, w)$ trained with Proximal Policy Optimisation (PPO)~\cite{schulman2017proximal} for $C$ iterations, instead of a Q-network $Q_{\theta}(s, a, \textbf{w})$. 

We adopt a multi-policy approach, each $\pi_{\theta}(s, w)$ specializes in a sub-space of $\textbf{W}$, enabling time-efficient and parallel optimisation with a compact parameter size. Inspired by~\cite{Liu2021PredictionGM}, we frame HV maximisation as a surrogate-assisted optimisation problem, where a data-driven surrogate predicts changes in objective values to select weight vectors that improve HV. We extend~\cite{Liu2021PredictionGM} by incorporating prediction uncertainty into the UCB acquisition function, balancing exploration and exploitation. Further, we replace large Pareto archives with scalarisation-vector-conditioned policies $\pi_{\theta}(s, w)$, reducing memory overhead while ensuring efficient PF coverage.

\vspace{1em} 
\noindent In summary, two contributions distinguish our work from previous systems of MORL/D:
\begin{enumerate}
\item A two-layer decomposition enables policies to specialize in different subspaces of the scalarisation vector space, with conditioning refining sub-problems within each subspace. This facilitates generalisation across local scalarisation neighbourhoods, enhancing HV metrics~\cite{Felten2023ATF}.

\item The use of a UCB acquisition function for selecting from a finite set of evenly distributed scalarisation vectors to balance exploration and exploitation.
\end{enumerate}

The rest of the paper is structured as follows. Background knowledge is introduced in Section \ref{sec:Preliminaries}. Prior work on gradient-based MORL is reviewed in Section \ref{sec:related_work}. The proposed method is described in Section \ref{sec:methodology}. Experiment configurations on six multi-objective benchmark problems are outlined in Section \ref{sec:experiments}, and SOTA results on these problems are discussed in Section \ref{sec:results}. Finally, we conclude in Section \ref{sec:conclusion} with directions for future research.

%%%%%%%%%%%%%%%%%%%%%%%%%%%%%%%%%%%%%%%%%%%%%%%%%%%%%%%%%%%%%%%%%%%%%%%%%%%%%%%%%%%%%%%%%%%%%%%%%%%%%%%%%%%%%%%%%%%%%%%%%%%%%%%%%%%%%%%%%%%%%%%%

%%%%%%%%%%%%%%%%%%%%%%%%%%%%%%%%%%%%%%%%%%%%%%%%%%%%%%%%%%%%%%%%%%%%%%%%%%%%%%%%%%%%%%%%%%%%%%%%%%%%%%%%%%%%%%%%%%%%%%%%%%%%%%%%%%%%%%%%%%%%%%%%

\section{Preliminaries}
\label{sec:Preliminaries}

\subsection{Policy-gradient Reinforcement Learning}
Policy gradient methods are a class of RL algorithms that optimise policies directly by computing gradients of an objective function with respect to the policy parameters. The objective function $J(\theta)$ for a parameterised policy \( \pi_\theta(a|s) \) can be expressed as the expected return, the policy gradient is then computed as:

\[
\nabla_\theta J(\theta) = \mathbb{E}_{\tau \sim \pi_\theta} \left[ \nabla_\theta \log \pi_\theta(a_t|s_t) \hat{A}_t \right],
\]

\noindent where \( \hat{A}_t \) is the advantage function \cite{sutton2000policy}, which measures the relative value of taking action \( a_t \) in state \( s_t \).

Proximal Policy Optimisation (PPO) \cite{schulman2017proximal} improves upon traditional policy gradient methods by introducing a surrogate objective that prevents overly large policy updates, ensuring stable learning. The PPO objective is defined as:

\[
L^{\text{PPO}}(\theta) = \mathbb{E}_t \left[ \min \left( r_t(\theta) \hat{A}_t, \text{clip}(r_t(\theta), 1-\epsilon, 1+\epsilon) \hat{A}_t \right) \right],
\]

\noindent where \( r_t(\theta) = \frac{\pi_\theta(a_t|s_t)}{\pi_{\theta_{\text{old}}}(a_t|s_t)} \) is the probability ratio, and \( \epsilon \) is a hyperparameter controlling the clipping range. By clipping the ratio, PPO limits deviations from the current policy, leading to improved stability and sample efficiency. These advancements have made PPO one of the most widely used algorithms for continuous and high-dimensional control tasks.

\subsection{Definition of MORL}
A multi-objective sequential decision making problem can be formulated as a multi-objective Markov Decision Process (MOMDP) defined by the tuple $\langle S, A, P, \gamma, \rho_{0}, \textbf{r}\rangle$ with state space $S$, action space $A$, state transition probability $P : S \times A \times S \rightarrow [0,1]$, discount factor $\gamma$, initial state distribution $\rho_{0}$, and vector-valued reward function $\textbf{r} : S \times A  \rightarrow \mathbb{R}^m$ specifying one-step reward for each of the $m$ objectives. A decision-making policy $\pi : S \rightarrow A$ maps states into actions, for which a vector-valued value function $\textbf{V}^{\pi}$ is defined as:

\begin{equation}
    \textbf{V}^{\pi} = \mathbb{E} \Bigg[ \sum_{t=0}^{H} \gamma^{t} \textbf{r}(s_t, \alpha_t) | s_0 \sim \rho_0,  a_t \sim \pi \Bigg]
\end{equation}

\noindent where $H$ is the length of the horizon.

The space of utility functions for a MORL problem is typically populated by linear and non-linear functions of $\textbf{V}^{\pi}$. In this work we focus on the highly prevalent case where the utility functions are linear, taking the form of $u(\textbf{V}^{\pi}) = \textbf{w}^{\top}\textbf{V}^{\pi}$, where scalarisation vector \textbf{w} provides the parametrisation of the utility function. Each element of $\textbf{w} \in \mathbb{R}^m$  specifies the relative importance (preference) of each objective. The space of linear scalarisation vectors \textbf{W} is the m-dimensional simplex: $\sum_i w_i=1, w_i \geq 0, i=1,\ldots,m$. For any given \textbf{w}, the original MOMDP is reduced to a single-objective MDP.

\subsection{Convex Coverage Set (CCS)}
Linear utility functions enable MORL to generate a Convex Coverage Set (CCS) of policies, which is a subset of all possible policies such as there exists a policy $\pi$ in the set that is optimal with respect to any linear scalarisation vector \textbf{w}:

\begin{equation}
    \mathit{CCS} \equiv \left\{ \textbf{V}^{\pi} \in \Pi \; | \; \exists \; \textbf{w} \; s.t. \; \forall \; \textbf{V}^{\pi'} \in \Pi, \textbf{V}^{\pi} \textbf{w}^{\top} \geq \textbf{V}^{\pi'} \textbf{w}^{\top}     \right\}
\end{equation}

\noindent CCS is a subset of PF defined for monotonically increasing utility functions~\cite{hayes2022practical}. At the same time, the underlying linear utility function space (space of scalarisation vectors \textbf{W}) is easier to search than that of arbitrarily-composed utility functions, where one has to search for the overall function composition out of primitive function elements. The sampling of scalarisation vectors $\textbf{w} \in \textbf{W}$ impacts the quality of the CCS in approximating the PF. Vectors can be sampled uniformly at random~\cite{abels2019dynamic,chenxi_metalearning,Lu2023MultiObjectiveRL,Natarajan2005DynamicPI,Yang2019AGA}, selected from a predefined set~\cite{basaklar2022pd,Liu2021PredictionGM}, or adapted via search methods~\cite{alegre2023sample,mossalam2016multiobjective}. A small step size is required for high-quality CCS when using predefined vectors~\cite{Roijers2015PointBasedPF}.

%%%%%%%%%%%%%%%%%%%%%%%%%%%%%%%%%%%%%%%%%%%%%%%%%%%%%%%%%%%%%%%%%%%%%%%%%%%%%%%%%%%%%%%%%%%%%%%%%%%%%%%%%%%%%%%%%%%%%%%%%%%%%%%%%%%%%%%%%%%%%%%%

%%%%%%%%%%%%%%%%%%%%%%%%%%%%%%%%%%%%%%%%%%%%%%%%%%%%%%%%%%%%%%%%%%%%%%%%%%%%%%%%%%%%%%%%%%%%%%%%%%%%%%%%%%%%%%%%%%%%%%%%%%%%%%%%%%%%%%%%%%%%%%%%

\section{Related Work}
\label{sec:related_work}
A thorough review of MORL can be found in~\cite{hayes2022practical}. In this section we highlight previous work that employs gradient-based with a linear utility function, which can be classified into three main categories. 

In the first category, a single policy ~\cite{pan2020additional,siddique2020learning} is trained using a linear scalarisation function when the user's preferences are known in advance. For example, ~\cite{basaklar2022pd} proposes a novel algorithm that trains a single universal network to cover the entire preference space. The approach utilises preferences to guide network parameter updates and employs a novel parallelisation strategy to enhance sample efficiency. 

In the second category, when user preferences are unknown or difficult to define, a CCS of policies is computed by training multiple independent policies with different scalarisation vectors to capture various trade-offs between objectives ~\cite{Li2019DeepRL,Lu2023MultiObjectiveRL}. The work of~\cite{mossalam2016multiobjective} proposes Optimistic Linear Support, a method that adaptively selects the weights of the linear utility function via the concepts of corner weights and estimated improvement to prioritise those corner weights. ~\cite{alegre2023sample} introduces a sample-efficient MORL algorithm that leverages Generalised Policy Improvement (GPI) to prioritize training on specific preference weights. By focusing on corner weights with higher GPI priority, the method iteratively learns a set of policies whose value vectors approximate the CCS. The work of~\cite{Liu2021PredictionGM} maintains a Pareto archive of policies by focusing on those scalarisation vectors that are expected to improve the hypervolume and sparsity metrics of the resulting PF the most. 

The third category of methods maintains a single policy conditioned on the scalarisation vector $\textbf{w}$. Several works aim to train such policies for few-shot adaptation to varying objective preferences~\cite{abels2019dynamic,chenxi_metalearning,Yang2019AGA}. \cite{abels2019dynamic} uses a scalarisation-vector-conditioned Q-network trained on randomly sampled $\textbf{w}$ to solve single-objective RL sub-problems, generalizing across changing objective preferences. Similarly, \cite{Yang2019AGA} applies Envelope Q-learning to train a Q-network for few-shot adaptation to new scalarisation vectors. \cite{chenxi_metalearning} proposes a PPO-based meta-policy trained collaboratively with data from policies specialized to sampled scalarisation vectors, enabling PF construction via few-shot fine-tuning. Other works directly approximate the CCS using a single scalarisation-conditioned policy. \cite{Lu2023MultiObjectiveRL} enhances Soft Actor Critic with an entropy term to train scalarisation-conditioned policies and Q-networks, while \cite{basaklar2022pd} extends~\cite{Liu2021PredictionGM} by incorporating scalarisation-conditioning, replacing a Pareto archive with a single policy.

Existing methods demonstrate notable strengths but face challenges in scalability, computational cost, generalisation, and performance consistency. By comparing these methods to UCB-MOPPO, we highlight its key advantages: 1) Preference space decomposition enables scalable, efficient parallel training with compact policies. 2) The UCB-driven surrogate model selects scalarisation vectors to maximise hypervolume. 3) Integration of PPO enhances stability and sample efficiency. 4) The weight-conditioned network reduces memory overhead.

%%%%%%%%%%%%%%%%%%%%%%%%%%%%%%%%%%%%%%%%%%%%%%%%%%%%%%%%%%%%%%%%%%%%%%%%%%%%%%%%%%%%%%%%%%%%%%%%%%%%%%%%%%%%%%%%%%%%%%%%%%%%%%%%%%%%%%%%%%%%%%%%

%%%%%%%%%%%%%%%%%%%%%%%%%%%%%%%%%%%%%%%%%%%%%%%%%%%%%%%%%%%%%%%%%%%%%%%%%%%%%%%%%%%%%%%%%%%%%%%%%%%%%%%%%%%%%%%%%%%%%%%%%%%%%%%%%%%%%%%%%%%%%%%%
\section{Methods}
\label{sec:methodology}

The key technical characteristics of the proposed MORL/D algorithm, named UCB-MOPPO, are as follows:

\textbf{MORL/D as scalar RL sub-problems.} 
The overall scalarisation weight \textbf{W} is divided into $K$ sub-spaces as shown in Figure \ref{fig:scalarisation_vector_decomposition}. A separate policy $\pi_k$ is trained for each sub-problem, conditioned on scalarisation vectors sampled from the corresponding sub-space $\textbf{W}_k \subset \textbf{W}$, where $k=1,\ldots,K$.

\textbf{Scalarisation-vector-conditioned Actor-Critic.} 
A policy network $\pi_{\theta}$, a value network $v^{\pi}_{\phi}$, and a scalarisation vector \textbf{w} $\left(\sum_i w_i=1, w_i \geq 0, i=1,\ldots,m \right)$ are used to maximise the weighted-sum of expected rewards, denoted as $J(\theta, \phi, \textbf{w}) = \sum_{i=1}^m w_i V_i^{\pi}$. Following~\cite{abels2019dynamic,Lu2023MultiObjectiveRL,Yang2019AGA}, both $\pi_{\theta}(s,\textbf{w})$ and $v^{\pi}_{\phi}(s,\textbf{w})$ are conditioned on \textbf{w}. This enables a single policy to express different trade-off between objectives by generalising across a neighbourhood of scalarisation vectors.

\textbf{Surrogate-assisted maximisation of CCS hypervolume.} 
An acquisition function based on UCB~\cite{srinivas2009gaussian} is used to select scalarisation vectors for training from each sub-space $\mathbf{W}_k$. During training, the chosen scalarisation vectors are those expected to maximise the hypervolume of the resulting CCS most effectively.

\subsection{Two-layer decomposition of MORL problem into scalar RL sub-problems}

At the first layer of problem decomposition, a set of $K$ evenly distributed scalarisation vectors, named as \emph{pivots}, are defined within the scalarisation vector space \textbf{W}. These pivot vectors effectively divide \textbf{W} into $K$ different sub-spaces as shown in Figure \ref{fig:scalarisation_vector_decomposition}. The system allows to independently train multiple pivot policies with different random seeds in order to improve the density of the resulting CCS. At the second layer of problem decomposition, for each sub-space $\textbf{W}_k$, $k=1,\ldots,K$, a number of $M$ evenly distributed scalarisation vectors are defined in turn. Therefore the overall MORL problem is decomposed into a total of $K*M$ scalar RL sub-problems, defined as follows for a fixed \textbf{w}:

\begin{equation}
\pi(\cdot, \textbf{w}) = \operatorname*{argmax}_{\pi'(\cdot, \textbf{w})} \mathbb{E} \left[ \sum_{t=1}^{H} \gamma^t \textbf{w}^{\top} \textbf{r}( s_t, \alpha_t) \right]
\end{equation}

\noindent A separate policy $\pi_k$ (i.e., \emph{pivot} policy) is independently trained for each sub-space $\textbf{W}_k \subset \textbf{W}$ by conditioning on the corresponding scalarisation vectors. As an example, for a bi-objective problem with $\textbf{w}=[w_1, w_2]$ in Figure~\ref{fig:scalarisation_vector_decomposition}, for two sub-spaces $\textbf{W}_i$, $\textbf{W}_j$ with $i>j$, $\left(w_1^{i,n} > w_1^{j,n}\right)$ $\land$ $\left(w_2^{i,n} < w_2^{j,n}\right)$, $n=1,\ldots,M$. The solutions of the $K*M$ sub-problems compose a CCS.

\subsection{Scalarisation-vector-conditioned Actor-Critic}

\begin{figure}[ht]
\begin{center}
\centerline{\includegraphics[width=0.42\columnwidth]{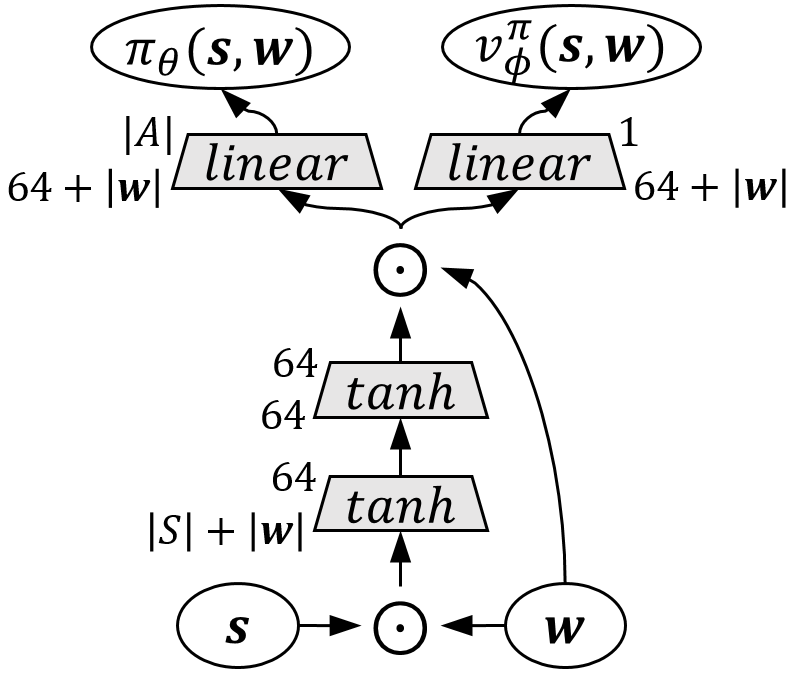}}
\caption{Scalarisation vector $\mathbf{w}$ conditioned actor-critic network.}
\label{fig:nn}
\end{center}
\end{figure}

The neural network(NN) architecture of the Actor-Critic is illustrated in Figure~\ref{fig:nn}. The state vector \textbf{s} concatenated with the scalarisation vector \textbf{w} form the input layer. \textbf{w} is also concatenated with the last shared layer through a residual connection~\cite{he2016deep}. Residual connections aim at improving the sensitivity of NN output to changes in \textbf{w} in a similar vein with reward-conditioned policies in~\cite{kumar2019reward}. 

\subsection{Surrogate-assisted maximisation of CCS hypervolume}

\subsubsection{Overview}
\label{sec:overview}
The learning algorithm proceeds in stages of $C$ iterations each. At every stage, $K$ pivot policies $\pi_k$, $k=1,\ldots,K$ are trained in parallel using PPO, each conditioned on a subset of $N$ scalarisation vectors out of $M$ in total that are defined for each corresponding sub-space $\textbf{W}_k$, $k=1,\ldots,K$, with $N \ll M$. The selection of $N$ scalarisation vectors to train on at each stage is performed via surrogate-assisted maximisation of the hypervolume of the CCS. First, a data-driven uncertainty-aware surrogate model is built in the scalarisation vector space to predict the expected change in each objective after training $\pi_k$ conditioned on said scalarisation vectors for $C$ iterations. Second, an acquisition function is defined as the UCB of the CCS hypervolume that is expected by including a scalarisation vector to the policy's conditioning set at each training stage. Maximising the acquisition function selects the scalarisation vectors that are expected to improve CCS hypervolume the most. Pseudo code can is provided in supplementary material, section 1, algorithm 3.

\subsubsection{Warm-up phase}

The algorithm starts with a warm-up stage performed using Algorithm \ref{alg:fixweight}. For an $m$-objective problem, a set of $K$ evenly distributed pivot vectors $\{\textbf{w}_i\}_{i=1}^K$ are generated, where $\sum_i w_i=1, w_i \geq 0, i=1,\ldots,m $. Accordingly, a set of $K$ pivot policies, each conditioned on a separate pivot vector, is trained using Algorithm ~\ref{alg:fixweight} for a number of epochs. 

\subsubsection{Surrogate model}

Let $\pi_{k,z+1}$ be the policy that results from the $k^{th}$ policy $\pi_{k,z}$ during the $z^{th}$ training stage of $C$ iterations. Let $\Delta V^{\pi_k, (z \rightarrow z+1)}_{j,\textbf{w}} = V^{\pi_{k,z+1}}_{j,\textbf{w}} - V^{\pi_{k,z}}_{j,\textbf{w}}$ be the change in the value of the $j^{th}$ objective for the $k^{th}$ policy conditioned on scalarisation vector \textbf{w} trained with PPO for $C$ iterations. For each pivot policy $\pi_k$, $k = 1, \ldots, K$, and for each objective $j=1,\ldots,m$, a separate dataset $D^{k,j}_{surrogate}$ is created using policy's $\pi_k$ evaluation data of objective $j$ that are collected from the simulation environment during a number of consecutive training stages $Z$, as follows:

\begin{equation}
D^{k,j}_{surrogate} = \Bigg\{ \left(\textbf{w}, \left(V^{\pi_{k,z+1}}_{j,\textbf{w}} - V^{\pi_{k,z}}_{j,\textbf{w}}\right) \right)\Bigg\} _{z=1}^{Z}
\end{equation}

\noindent A surrogate model $f^{k,j}_{bagging} : \mathbb{R}^{m} \rightarrow \mathbb{R}$ is trained on $D^{k,j}_{surrogate}$ to predict $\Delta V^{\pi_k, (z \rightarrow z+1)}_{j,\textbf{w}}$ as a function of the scalarisation vector \textbf{w}, for $m$ number of objectives. The training is incremental while additional tuples are appended to $D_{surrogate}$ from consecutive training stages. The surrogate model takes the form of Bagging~\cite{Breiman1996BaggingP} of linear models $f_{\psi}(\textbf{w}) = \sum_{i=1}^m \psi_i w_i + \psi_0$ with elastic net regularisation~\cite{hastieElasticNet}. Bagging trains independently $B$ linear models $\{f^b_\psi\}_{b=1}^B$ on $B$ bootstrap samples of the original training data and predicts using their average, that is $f^{k,j}_{bagging}(\textbf{w}) = \frac{1}{B}\sum_{i=1}^B f^b_\psi (\textbf{w})$. An estimate of the epistemic uncertainty of the prediction can be computed using the variance of the component model predictions~\cite{gawlikowski2023survey}, that is $\sigma^2_{k,j}(\textbf{w}) = \frac{1}{B} \sum_{i=1}^B \left(f^b_\psi (\textbf{w}) - f^{k,j}_{bagging}(\textbf{w}) \right)^2$. 

\subsubsection{UCB acquisition function maximisation}
\begin{figure*}[ht]
\centering % Ensure the entire figure is centered

% Left Subfigure
\begin{subfigure}{0.45\textwidth}
    \centering
    \includegraphics[width=0.80\linewidth]{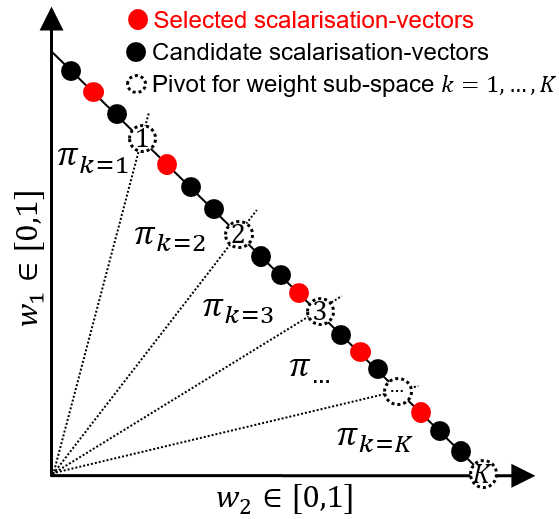}
    \caption{Decomposition of MORL problem into scalar sub-problems.}
    \label{fig:scalarisation_vector_decomposition}
\end{subfigure}
% Adjust spacing as necessary
\hfill
% Right Subfigure
\begin{subfigure}{0.45\textwidth}
    \centering
    \includegraphics[width=0.882\linewidth]{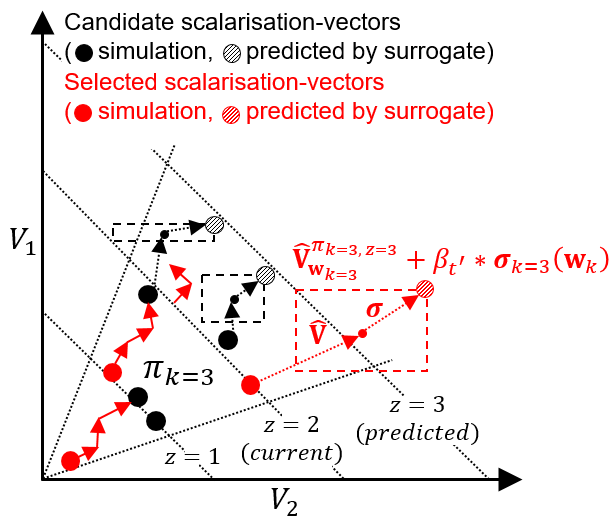}
    \caption{Surrogate-assisted maximisation of CCS hypervolume.}
    \label{fig:ucb_driven_search}
\end{subfigure}

\caption{Overview of the proposed approach for UCB-driven utility function search for MORL/D.}
\label{fig:system_diagram}
\end{figure*}

For each policy $k$ and each objective $j$ a surrogate model can predict the expected objective values by conditioning a policy on scalarisation vector \textbf{w} during a training stage $z$ as follows:

\begin{equation}
\hat{V}^{\pi_{k,z+1}}_{j,\textbf{w}} = V^{\pi_{k,z}}_{j,\textbf{w}} + f^{k,j}_{bagging}(\textbf{w}) 
\end{equation}

\noindent with a corresponding vectorised prediction, denoted as  $\hat{\textbf{V}}^{\pi_{k,z+1}}_{\textbf{w}} = \textbf{V}^{\pi_{k,z}}_{\textbf{w}} + \textbf{f}^{k}_{bagging}(\textbf{w})$ over $j$ objectives and the accompanying vectorised uncertainty estimate $\boldsymbol{\sigma}^2_{k}(\textbf{w})$.

At the beginning of each training stage $z$ the algorithm needs to select those $N$ scalarisation vectors out $M$ evenly distributed vectors in each sub-space $\textbf{W}_k$, $k = 1, \ldots, K$ that are predicted via the surrogate model to improve the hypervolume of the resulting $\texttt{HV}(\mathit{CCS})$ the most. $\texttt{HV}(\cdot)$ is a function that computes hypervolume as in~\cite{Felten2023ATF}. The scalarisation vectors are selected one at a time without replacement in a sequence of $N$ invocations of a process that maximises a UCB acquisition function defined on $\texttt{HV}(\mathit{CCS})$ as:

% \begin{align*}
% % \label{eq:acquisition_func}
% \textbf{w}_k^* &= \argmax_{\lbrace\textbf{w}_{k,j} \rbrace_{j=1}^M} \texttt{HV} \left (\texttt{Pareto}\left(\hat{\textbf{V}}^{\pi_{k,z+1}}_{\textbf{w}_{k,j}} + \beta_{t'}*\boldsymbol{\sigma}_{k}(\textbf{w}_{k,j}) \right) \right) \numberthis \label{eq:acquisition_func}
% \end{align*}

\begin{align}
\textbf{w}_k^* &= \operatorname*{argmax}_{\textbf{w}_{k,j} \in \lbrace \textbf{w}_{k,j} \rbrace_{j=1}^M} 
\texttt{HV} \big( \texttt{Pareto}( \textbf{V}_{k,j} ) \big), \label{eq:acquisition_func} \\
\textbf{V}_{k,j} &= \hat{\textbf{V}}^{\pi_{k,z+1}}_{\textbf{w}_{k,j}} 
+ \beta_{t'} \cdot \boldsymbol{\sigma}_{k}(\textbf{w}_{k,j}), \quad k = 1, \ldots, K. \nonumber
\end{align}

\noindent where  $\texttt{Pareto}(\cdot)$ is a function computing the CCS from a set of objective vectors. The dynamic parameter of current training step,i.e.,  $\beta_{t'} = \sqrt{\frac{log(2t')}{t'}}$ ensures that scalarisation vector with higher uncertainty (standard deviation) are explored early, but the focus shifts towards exploiting the ones with high mean rewards as confidence improves, assuming maximisation of objectives. Once selected, a scalarisation vector $\textbf{w}_{k}^{*}$ is removed from the candidate set $\{ \textbf{w}_{k,j} \}_{j=1}^M$ for the current training stage. The process of selecting scalarisation vectors that are expected to maximise the hypervolume of the resulting CCS is illustrated in Figure~\ref{fig:ucb_driven_search}.

\subsection{Baseline Methods}
\label{sec:baselines}

Seven baseline methods are introduced for comparison, including three proposed \textbf{MOPPO} baseline methods for ablation studies and four SOTA MORL methods: 1) \textbf{PG-MORL~\cite{Liu2021PredictionGM}}, 2) \textbf{PD-MORL~\cite{basaklar2022pd}}, 3) \textbf{CAPQL~\cite{Lu2023MultiObjectiveRL}}, and 4) \textbf{GPI-LS~\cite{alegre2023sample}}. These baselines are used to evaluate the benefits introduced by PPO and the UCB-driven search of the scalarisation vector space. A brief description of the proposed \textbf{MOPPO baselines} is as follows:

\begin{algorithm}[ht]
   \caption{Fixed-MOPPO}
   \label{alg:fixweight}
   \begin{algorithmic}[1]
   \State \textbf{Input:} State $s_t$, weights $\mathbf{w}$
   \State \textbf{Initialize:} $K$ Actor-Critic networks $\pi_k$, $v^{\pi_k}$; scalarisation spaces $\mathbf{W}_k$; buffer $\mathcal{E}$ of size $D$.

   \For{$k = 1$ \textbf{to} $K$}  
       \For{$t = 1$ \textbf{to} $D$}
           \State $\mathbf{w}_{\text{pivot}} \gets \text{get\_pivot\_weight}(\mathbf{W}_k)$
           \State $a_t \sim \pi_k(s_t, \mathbf{w}_{\text{pivot}})$, $s_{t+1}, \mathbf{r}_t \gets \text{simulator}(s_t, a_t)$
           \State $\mathcal{E} \gets \mathcal{E} \cup \langle s_t, a_t, \mathbf{w}_{\text{pivot}}, \mathbf{r}_t, s_{t+1} \rangle$, $s_t \gets s_{t+1}$
       \EndFor
       \State Sample $\langle s_t, a_t, \mathbf{w}_{\text{pivot}}, \mathbf{r}_t, s_{t+1} \rangle \sim \mathcal{E}$
       \State $\theta \gets \theta + \eta \nabla_\theta \log \pi_\theta(s_t, a_t; \mathbf{w}_{\text{pivot}}) A^\pi(s_t, a_t; \mathbf{w}_{\text{pivot}})$
       \State $\phi \gets \phi + \eta \| V^{\pi_k}(s_t; \mathbf{w}_{\text{pivot}}) - (\mathbf{r}_t + \gamma V^{\pi_k}(s_{t+1}; \mathbf{w}_{\text{pivot}})) \|^2$
       \State $\mathcal{E} \gets \emptyset$
   \EndFor
   \end{algorithmic}
\end{algorithm}

\textbf{Fixed-MOPPO} in Algorithm~\ref{alg:fixweight}, which trains $K$ policies, each conditioned on a fixed scalarisation vector $\mathbf{w}_\mathit{pivot}\in\mathbf{W}_k$ corresponding to subspace $k$ (see Figure~\ref{fig:scalarisation_vector_decomposition}). 

\textbf{Random-MOPPO} (see supplementary material, section 1, algorithm 2) which is similar to Fixed-MOPPO, $K$ policies are trained, with each $\pi_k$ conditioned on scalarisation vectors uniformly sampled from $\textbf{W}_k$. These vectors are periodically re-sampled to ensure diversity during training. 

\textbf{Mean-MOPPO} which is identical to UCB-MOPPO, except that the acquisition function ignores uncertainty by setting $\beta_{t'} := 0, \forall t'$ in Equation~\ref{eq:acquisition_func}.

%%%%%%%%%%%%%%%%%%%%%%%%%%%%%%%%%%%%%%%%%%%%%%%%%%%%%%%%%%%%%%%%%%%%%%%%%%%%%%%%%%%%%%%%%%%%%%%%%%%%%%%%%%%%%%%%%%%%%%%%%%%%%%%%%%%%%%%%%%%%%%%%
%%%%%%%%%%%%%%%%%%%%%%%%%%%%%%%%%%%%%%%%%%%%%%%%%%%%%%%%%%%%%%%%%%%%%%%%%%%%%%%%%%%%%%%%%%%%%%%%%%%%%%%%%%%%%%%%%%%%%%%%%%%%%%%%%%%%%%%%%%%%%%%%

\section{Experiment Setup}

We evaluate the performance of all seven baselines on six continuous control multi-objective RL problems: \textbf{Swimmer-V2}, \textbf{Halfcheetah-V2}, \textbf{Walker2d-V2}, \textbf{Ant-V2}, \textbf{Hopper-V2}, and \textbf{Hopper-V3} (the only problem with three objectives). Detailed descriptions of the objectives, as well as the state and action spaces, are provided in the supplementary material, section 2. Each method is evaluated on three different random seeds for each problem, consistent with \cite{Liu2021PredictionGM}. For a fair comparison, we use raw reward outputs and assess the results using the \emph{hypervolume} (HV) and \emph{Expected Utility} (EU) metrics~\cite{alegre2023sample}.

\label{sec:experiments}
\setlength{\tabcolsep}{5pt} % Adjust column spacing
\begin{table}[ht]
    \caption{Decomposition setup for two-objective and three-objective problems.}
    \label{tab:decompositionParams}
    \centering
    \begin{tabular}{lcccccc}
        \toprule
        \textbf{Parameter} & \textbf{Objective} & \textbf{UCB} & \textbf{Mean} & \textbf{Random} & \textbf{Fixed} \\ 
        \midrule
        $K$ (pivot vectors) & 2-objective & 10 & 10 & 10 & 10 \\
                            & 3-objective & 36 & 36 & 36 & 36 \\ 
        \midrule
        $N$ (sub-space selection) & 2-objective & 10 & 10 & 10 & 1 \\
                                  & 3-objective & 10 & 10 & 10 & 1 \\ 
        \midrule
        $M$ (sub-space vectors) & 2-objective & 100 & 100 & 10 & 10 \\
                                & 3-objective & 117 & 117 & 36 & 36 \\ 
        \midrule
        Step-size (layer 1) & 2-objective & 0.1 & 0.1 & 0.1 & 0.1 \\
                            & 3-objective & 0.1 & 0.1 & 0.1 & 0.1 \\ 
        \midrule
        Step-size (layer 2) & 2-objective & 0.01 & 0.01 & 0.1 & 0.1 \\
                            & 3-objective & 0.05 & 0.05 & 0.1 & 0.1 \\ 
        \bottomrule
    \end{tabular}
\end{table}

For UCB-MOPPO and the proposed MOPPO baselines, hyperparameters for policy neural network initialisation and PPO are aligned with the values recommended in Stable-Baselines3 \cite{raffin2021stable}. Detailed hyperparameters are provided in the supplementary material, section 3. Proposed problem decomposition is based on scalarisation vectors generated via discretisation of space \textbf{W} with a fixed step size. The overall parameters of problem decomposition for 2- and 3-objective problems are summarised in Table~\ref{tab:decompositionParams}. SOTA baselines are reproduced using the MORL-Baseline library \cite{Felten2023ATF}. 

%%%%%%%%%%%%%%%%%%%%%%%%%%%%%%%%%%%%%%%%%%%%%%%%%%%%%%%%%%%%%%%%%%%%%%%%%%%%%%%%%%%%%%%%%%%%%%%%%%%%%%%%%%%%%%%%%%%%%%%%%%%%%%%%%%%%%%%%%%%%%%%%
%%%%%%%%%%%%%%%%%%%%%%%%%%%%%%%%%%%%%%%%%%%%%%%%%%%%%%%%%%%%%%%%%%%%%%%%%%%%%%%%%%%%%%%%%%%%%%%%%%%%%%%%%%%%%%%%%%%%%%%%%%%%%%%%%%%%%%%%%%%%%%%%

\section{Results Analysis}
\label{sec:results}

\subsection{Quality of Pareto Front}

\begin{figure}[ht]
\vspace{0.5cm}
\begin{center}
\centerline{\includegraphics[width=\textwidth]{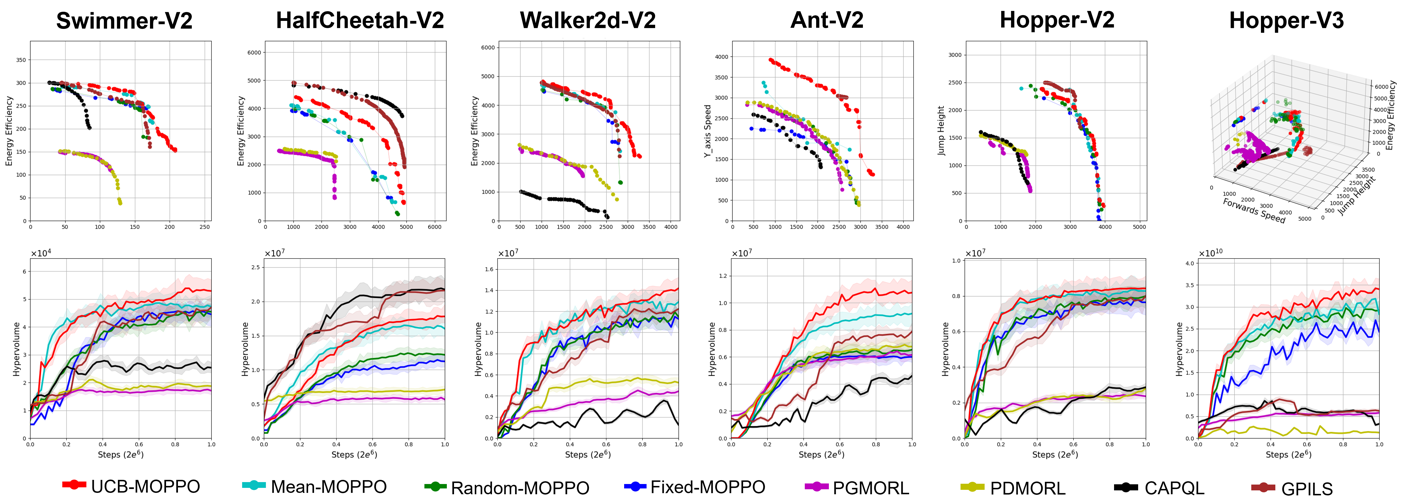}}
\caption{Comparison of PF and HV evaluation for six MORL tasks. The top row shows PF plots, and the bottom row shows HV comparisons. Shaded areas in HV plots represent the standard deviation over three seeds.}
\label{fig:combined_compare}
\end{center}
\end{figure}

Pareto front(PF) are evaluated in \textbf{HV and EU}, which are reported in Table~\ref{tb:hv_eu_compare}.
\textbf{PF coverage} and \textbf{HV convergence speed} are visualized in Figure~\ref{fig:combined_compare}.

\begin{table}[ht]
\caption{Evaluation of HV and EU metrics for continuous MORL tasks over three independent runs. Due to table size limitations,environment and method names are \textbf{abbreviated}, and only \textbf{mean} values from three runs are reported. The full detailed table is provided in supplementary material, section 4.}
\vspace{0.25cm}
\centering
\label{tb:hv_eu_compare}
\setlength{\tabcolsep}{4pt} % Adjust column spacing

\begin{tabular}{l l c c c c c c c c} 
% \textbf{Benchmark} & \textbf{Metric} & \textbf{UCB} & \textbf{Mean} & \textbf{Random} & \textbf{Fixed} & \textbf{~\cite{Liu2021PredictionGM}} & \textbf{~\cite{basaklar2022pd}} & \textbf{~\cite{Lu2023MultiObjectiveRL}} & \textbf{~\cite{alegre2023sample}} \\\hline

\textbf{} & \textbf{Metric} & \textbf{UCB} & \textbf{Mean} & \textbf{Rand-} & \textbf{Fix-} & \textbf{PG-} & \textbf{PD-} & \textbf{CAP-} & \textbf{GPI-} \\\hline

\multirow{2}{*}{Swimmer} & HV ($10^4$)   & \textbf{5.60} & 4.72 & 4.56 & 4.45 & 1.67 & 1.77 & 2.40 & 4.75 \\
                         & EU ($10^2$)   & \textbf{2.18} & 2.16 & 2.08 & 2.11 & 1.23 & 1.24 & 1.79 & 2.14 \\\hline
\multirow{2}{*}{Halfcheetah} & HV ($10^7$)   & 1.78 & 1.60 & 1.21 & 1.12 & 0.58 & 0.62 & \textbf{2.20} & 2.16 \\
                             & EU ($10^3$)   & 3.88 & 3.58 & 3.55 & 3.44 & 2.30 & 2.42 & \textbf{4.47} & 4.30 \\\hline
\multirow{2}{*}{Walker2d} & HV ($10^7$)   & \textbf{1.40} & 1.29 & 1.15 & 1.13 & 0.44 & 0.56 & 0.18 & 1.22 \\
                          & EU ($10^3$)   & \textbf{3.54} & 3.42 & 3.34 & 3.30 & 1.98 & 2.24 & 1.50 & 3.40 \\\hline
\multirow{2}{*}{Ant} & HV ($10^7$)   & \textbf{1.07} & 0.92 & 0.65 & 0.60 & 0.61 & 0.66 & 0.45 & 0.81 \\
                     & EU ($10^3$)   & \textbf{2.96} & 2.61 & 2.35 & 2.21 & 2.28 & 2.41 & 2.03 & 2.84 \\\hline
\multirow{2}{*}{Hopper2d} & HV ($10^7$)   & \textbf{0.84} & 0.81 & 0.79 & 0.76 & 0.23 & 0.25 & 0.25 & 0.80 \\
                          & EU ($10^3$)   & 2.77 & \textbf{2.84} & 2.77 & 2.71 & 1.48 & 1.51 & 1.46 & 2.75 \\\hline
\multirow{2}{*}{Hopper3d} & HV ($10^{10}$) & \textbf{3.62} & 2.83 & 2.90 & 2.64 & 0.63 & 0.11 & 0.31 & 0.65 \\
                          & EU ($10^3$)   & 3.20 & 3.19 & \textbf{3.28} & 2.90 & 1.70 & 1.74 & 1.52 & 2.14 \\\hline
\end{tabular}
\end{table}

\textbf{UCB-MOPPO} delivers consistent performance and outperforms all baselines except in Halfcheetah-V2. Notably, \textbf{PGMORL} produces a dense PF by accumulating a large number of policies in the archive, yet it achieves significantly lower HV and EU compared to the proposed MOPPO baselines. \textbf{PDMORL} achieves slightly higher HV and EU than PGMORL while using only a single network; however, it struggles with three-objective problems. \textbf{CAPQL} achieves the best performance in Halfcheetah-V2 but struggles significantly on Walker2d-V2 and Hopper-V3 due to learning only a limited dynamics of those problems. \textbf{GPI-LS} delivers the most competitive results and requires the longest GPU time for training. it performs better than the proposed MOPPO baselines in Halfcheetah-V2; however, it fails to explore a spread CCS in the Ant-V2 and Hopper-2d/-3d environments. 

The consistent performance of the proposed method can be partially attributed to the integration of PPO, which provides enhanced stability and sample efficiency by employing trust region enforcement compared to vanilla policy gradient methods. As shown in Figure~\ref{fig:combined_compare}, UCB-MOPPO demonstrates superior convergence across most tasks (except for HalfCheetah-V2). The proposed method achieves rapid early-stage convergence by efficiently identifying scalarisation vectors that improve hypervolume. Additionally, the MOPPO baselines exhibit stable convergence without overfitting and dropping, ensuring robust performance across diverse multi-objective tasks.

\subsection{Analysis of Ablation Experiments}

\textbf{Fixed-MOPPO} is the simplest baseline, conditioning policies on pivot scalarisation vectors. It provides quick insights into a new MORL problem but results in a sparse PF.

\textbf{Random-MOPPO} achieves higher hypervolume and a denser PF than Fixed-MOPPO by randomly selecting scalarisation vectors under sub-space during training. However, it is inefficient, failing to explore promising regions effectively.

\textbf{Mean-MOPPO} improves search efficiency but yields a less dense PF than UCB-driven methods. The less coverage of PF can be attributed to the greedy selection of scalarisation vectors based solely on the mean predicted hypervolume improvement from the surrogates, without awareness of the uncertainty.

\subsection{Comparison of the Policy Archive}
UCB-MOPPO constructs a satisfactory PF with a consistently low number of policies compared to the baselines across all tested environments. The total number of policies after convergence is shown in Table \ref{tb:rp}. Some challenging problems result in a large number of policies in the archive (e.g., Walker2d-V2 and Hopper-V3). 

\begin{table}[ht]
\centering
\caption{Number of archived policies after the hypervolume converges.}
\label{tb:rp}
\setlength{\tabcolsep}{5pt} % Adjust column spacing
\begin{tabular}{lcccc} 
\toprule
\textbf{Environment} & \textbf{PGMORL} & \textbf{CAPQL} & \textbf{GIPLS} & \textbf{UCB-MOPPO} \\ 
\midrule
Swimmer-v2    & 168  & 33  & 41  & 30  \\ 
Halfcheetah-v2 & 285  & 35  & 83  & 30  \\ 
Walker2d-v2    & 412  & 51  & 52  & 30  \\ 
Ant-v2         & 64   & 26  & 27  & 30  \\ 
Hopper-v2      & 206  & 30  & 45  & 30  \\ 
Hopper-v3      & 4023 & 82  & 89  & 108 \\ 
\bottomrule
\end{tabular}
\end{table}

\begin{figure}[ht]
\begin{center}
\centerline{\includegraphics[width=0.75\columnwidth]{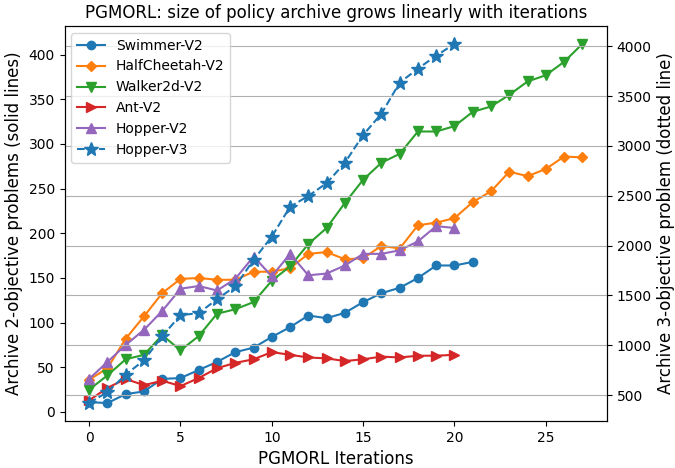}}
\caption{Growth in policy archive size in PGMORL.}
\label{fig:eps}
\end{center}
\end{figure}

A key advantage of all the MOPPO methods is that they maintain a small, fixed set of policies, unlike PGMORL, where the policy archive size grows linearly, as depicted in Figure \ref{fig:eps}. Therefore, an efficient search in UCB-MOPPO is complemented by a memory-efficient implementation, which can be an important consideration in production environments with tight resource constraints.

\vspace{-1.5cm}

\subsection{Interpolating in scalarisation vectors spaces}
An analysis was carried out to assess the ability of policies trained via UCB-MOPPO to interpolate in a more fine-grained discretisation of vector space $\mathbf{W}$, discretised using smaller step-sizes than those considered during training. 

%This observation aligns with our evaluation of policy efficiency, indicating a the positive effect of the weight condition approach.

\begin{figure}[ht]
\centering
% First Row
\begin{subfigure}{0.45\textwidth}
  \centering
  \includegraphics[width=0.9\linewidth]{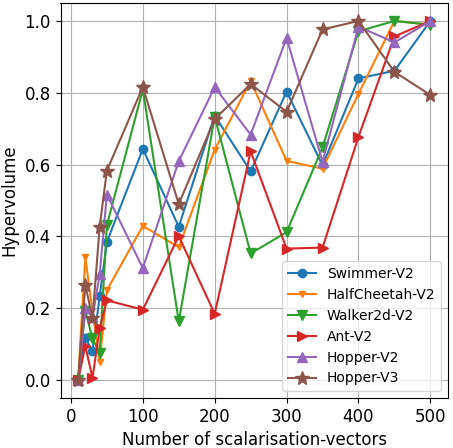}
  \label{fig:hvw}
\end{subfigure}
\begin{subfigure}{0.45\textwidth}
  \centering
  \includegraphics[width=0.9\linewidth]{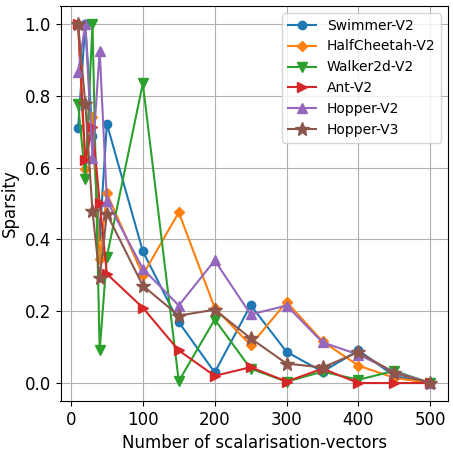}
  \label{fig:spw}
\end{subfigure}

\caption{Hypervolume and sparsity achieved by UCB-MOPPO improve with more scalarisation vectors conditioning the policies.}
\label{fig:hvsp}
% \vskip 0.1in
\end{figure}

Each of the \( K \) policies, corresponding to the \( K \) sub-spaces, is evaluated on \( N = \{10, 20, \dots, 50\} \cup \{100, 150, \dots, 500\} \) scalarisation vectors. The results are shown in Figure~\ref{fig:hvsp}, where we present the \texttt{hypervolume} and \texttt{sparsity} curves. The plots reveal a general trend: as \( N \) increases, the hypervolume improves, indicating a better approximation of the PF. Simultaneously, sparsity decreases, signifying a more uniform PF coverage. This suggests that increasing the granularity of scalarisation vectors leads to a more comprehensive and higher-quality representation of the PF, all without requiring additional training. This advantage is not shared by PGMORL or other MORL/D approaches, which often need further training to achieve similar results. Thus, the proposed method offers a distinct benefit in improving PF quality with minimal computational overhead.

%%%%%%%%%%%%%%%%%%%%%%%%%%%%%%%%%%%%%%%%%%%%%%%%%%%%%%%%%%%%%%%%%%%%%%%%%%%%%%%%%%%%%%%%%%%%%%%%%%%%%%%%%%%%%%%%%%%%%%%%%%%%%%%%%%%%%%%%%%%%%%%%
%%%%%%%%%%%%%%%%%%%%%%%%%%%%%%%%%%%%%%%%%%%%%%%%%%%%%%%%%%%%%%%%%%%%%%%%%%%%%%%%%%%%%%%%%%%%%%%%%%%%%%%%%%%%%%%%%%%%%%%%%%%%%%%%%%%%%%%%%%%%%%%%

\section{Conclusion}
\label{sec:conclusion}

This paper presents a method for efficiently searching scalarisation vectors that maximize the quality of the CCS. The key findings are: 
(1) the proposed method outperforms competitive baselines in terms of CCS hypervolume and Expected Utility in most cases, 
(2) it requires maintaining a minimal number of policies to produce high-quality CCS, making it well-suited for resource-constrained environments, and 
(3) the CCS hypervolume and sparsity metrics improve as the scalarisation vector step-size decreases, demonstrating effective generalisation across scalarisation vector neighbourhoods.
For future work, we plan to explore acquisition functions that leverage additional Pareto front quality indicators~\cite{falcon2020indicator} and search algorithms designed for non-linear utility function spaces~\cite{reymond2023actor}.

% \begin{credits}
% \subsubsection{\ackname} A bold run-in heading in small font size at the end of the paper is
% used for general acknowledgments, for example: This study was funded

% \subsubsection{\discintname}
% It is now necessary to declare any competing interests or to specifically
% state that the authors have no competing interests. Please place the
% statement with a bold run-in heading in small font size beneath the
% (optional) acknowledgments,
% for example: The authors have no competing interests to declare that are
% relevant to the content of this article. Or: Author A has received research
% grants from Company W. Author B has received a speaker honorarium from
% Company X and owns stock in Company Y. Author C is a member of committee Z.
% \end{credits}
\bibliographystyle{splncs04}
\bibliography{references}

\clearpage 
% Start appendix
% \appendix

\begin{center}
  \LARGE\bfseries Supplementary Material
\end{center}
\vspace{1em}

\addcontentsline{toc}{section}{Appendix}  % Optional: adds to TOC

% Reset section numbering for Appendix A, B, ...
\renewcommand{\thesection}{\Alph{section}}

\section{Algorithms}

\begin{algorithm}[ht]
   \caption{Fixed-MOPPO}
   \label{alg:fixweight}
   \begin{algorithmic}[1] % Enable line numbering
   \State \textbf{Input:} State $s_t$, weights $\textbf{w}$
   \State \textbf{Initialize:} $K$ weight-conditioned Actor-Critic Networks $\pi_{k}$ / $v^{\pi_{k}}$,  scalarisation-vector sub-spaces $\textbf{W}_k$ for $k=1,\ldots,K$, and memory buffer $\mathcal{E}$ size of $D$.
   \For{$k = 1$ \textbf{to} $K$}  
       \For{$t = 1$ \textbf{to} $D$}
            \State $\textbf{w}_\mathit{pivot} \leftarrow \text{get pivot weight}(\mathbf{W}_k)$       
           \State $a_{t} \leftarrow \pi_k(s_t,\mathbf{w}_\mathit{pivot})$
           \State $s_{t+1}, \mathbf{r_t} \leftarrow simulator(a_{t})$
           \State $\mathcal{E}  \leftarrow \mathcal{E} \cup \langle s_t, a_t, \mathbf{w}_\mathit{pivot}, \mathbf{r_t}, s_{t+1}\rangle$
           \State $s_t \leftarrow s_{t+1}$
       \EndFor
    
       \State sample $ \langle s_t, a_t, \mathbf{w}_\mathit{pivot}, \mathbf{r_t}, s_{t+1}\rangle \leftarrow \mathcal{E} $
       \State $\theta \leftarrow \theta + \eta \left(\nabla_\theta \log \pi_\theta (s_t, a; \mathbf{w}_\mathit{pivot})\right)\left(A^{\pi}(s_t, a_t; \mathbf{w}_\mathit{pivot})\right)$
       \State $\phi \leftarrow \phi + ||\boldsymbol{V}^{\pi_k}(s_t; \boldsymbol{\mathbf{w}_\mathit{pivot}}) - \boldsymbol{V}^{\pi_k}(s_{t+1}; \boldsymbol{\mathbf{w}_\mathit{pivot}})||^2$
       \State clear $\mathcal{E}$
   \EndFor
\end{algorithmic}
\end{algorithm}

\begin{algorithm}[ht]
   \caption{Random-MOPPO}
   \label{alg:randweight}
   \begin{algorithmic}[1] % Enable line numbering
   \State \textbf{Input:} State $s_t$, weights $\mathbf{w}$
   \State \textbf{Initialize:} $K$ weight-conditioned Actor-Critic Networks $\pi_{k}$ / $v^{\pi_{k}}$,  scalarisation-vector sub-spaces $\textbf{W}_k$ for $k=1,\ldots,K$, memory buffer $\mathcal{E}$ size of $D$, and scalarisation-vector re-sampling frequency $\mathit{RF}$.   
   \For{$k = 1$ \textbf{to} $K$}  
       \For{$t = 1$ \textbf{to} $D$}
           \If{$t\,\,\%\,\,\mathit{RF} = 0$} % Corrected modulus check
          \State $\mathbf{w}_t \leftarrow \text{uniform random sample}(\mathbf{W}_k)$
           \EndIf
        
           \State $a_{t} \leftarrow \pi_k(s_t,\boldsymbol{\mathbf{w}_t})$
           \State $s_{t+1}, \mathbf{r_t} \leftarrow simulator(a_{t})$
           \State $\mathcal{E}  \leftarrow \mathcal{E} \cup \langle s_t, a_t, \mathbf{w}_t, \mathbf{r_t}, s_{t+1}\rangle$
           \State $s_t \leftarrow s_{t+1}$
       \EndFor
    
       \State sample $ \langle s_t, a_t, \mathbf{w}_t, \mathbf{r_t}, s_{t+1}\rangle \leftarrow \mathcal{E} $
       \State $\theta \leftarrow \theta + \eta \left(\nabla_\theta \log \pi_\theta (s, a; \mathbf{w}_t)\right) \left(A^{\pi}(s_t, a_t; \mathbf{w}_t)\right)$
       \State $\phi \leftarrow \phi + ||\boldsymbol{V}^{\pi_k}(s_t; \mathbf{w}_t) - \boldsymbol{V}^{\pi_k}(s_{t+1}; \mathbf{w}_t)||^2$
       \State clear $\mathcal{E}$
   \EndFor
\end{algorithmic}
\end{algorithm}

\begin{algorithm}[!ht]
   \caption{UCB-MOPPO}
   \label{alg:ucb-moppo}
\begin{algorithmic}[1] \small
   \State \textbf{Input:} Environment state \(S_t\) and full weight set \(\mathbf{W}\).
   \State \textbf{Initialize:} 
       \(K\) weight-conditioned Actor-Critic networks \(\{(\pi_k, v^{\pi_k})\}_{k=1}^K\); for each \(k\), a predetermined subspace \(\mathbf{W}_k \subset \mathbf{W}\) (of size \(M\)); working pool \(\widetilde{\mathbf{W}}_k \gets \mathbf{W}_k\); number of objectives \(m\); warm-up iterations \(Q\); evaluation interval \(C\); for each \(k\) and \(j=1,\dots,m\), surrogate dataset \(D^{k,j}_{\text{surrogate}} \gets \emptyset\); surrogate models \(\{f^{k,j}_{\text{bagging}}\}\); and dynamic weight pool \(\mathcal{E}\) (of size \(D\)).
   
   \For{\(t = 0\) \textbf{to} \(T\)}
       \Comment{Warm-up Stage}
       \For{\(k = 1\) \textbf{to} \(K\)}
           \State \(\widetilde{\mathbf{W}}_k \gets \text{getPivotWeights}(\mathbf{W}_k)\)
           \State \(\pi_k \gets \text{FixWeightOptimization}(\pi_k, \widetilde{\mathbf{W}}_k)\)
       \EndFor
       
       \Comment{Periodic Evaluation (every \(C\) iterations)}
       \If{\(t \mod C = 0\)}
           \For{\(k = 1\) \textbf{to} \(K\)}
               \ForAll{\(\mathbf{w} \in \widetilde{\mathbf{W}}_k\)}
                   \State \(V^{\pi_k}_{j,\mathbf{w}} \gets \text{Simulate}(\pi_k, \mathbf{w})\)  \Comment{for each objective \(j\)}
               \EndFor
           \EndFor
       \EndIf
       
       \If{\(t > Q\)}
           \Comment{Construct Surrogate Training Data}
           \For{\(k = 1\) \textbf{to} \(K\)}
               \ForAll{\(\mathbf{w} \in \widetilde{\mathbf{W}}_k\)}
                   \For{\(z = 0\) \textbf{to} \(\lfloor t/C \rfloor - 1\)}
                       \For{\(j = 1\) \textbf{to} \(m\)}
                           \State \(\Delta V^{\pi_k,(z\to z+1)}_{j,\mathbf{w}} \gets V^{\pi_k,z+1}_{j,\mathbf{w}} - V^{\pi_k,z}_{j,\mathbf{w}}\)
                           \State Append \((\mathbf{w}, \Delta V^{\pi_k,(z\to z+1)}_{j,\mathbf{w}})\) to \(D^{k,j}_{\text{surrogate}}\)
                       \EndFor
                   \EndFor
               \EndFor
           \EndFor
           
           \Comment{Update Surrogate Models}
           \For{\(k = 1\) \textbf{to} \(K\)}
               \For{\(j = 1\) \textbf{to} \(m\)}
                   \State Update \(f^{k,j}_{\text{bagging}}\) using \(D^{k,j}_{\text{surrogate}}\)
               \EndFor
           \EndFor
           
           \Comment{Scalarisation-Vector Selection via UCB Acquisition}
           \For{\(k = 1\) \textbf{to} \(K\)}
               \State \(\mathcal{D}_k \gets \emptyset\)
               \ForAll{\(\mathbf{w} \in \mathbf{W}_k\)}  \Comment{Candidates from the full subspace}
                   \State \(V^{\pi_k}_{\mathbf{w}} \gets \text{Simulate}(\pi_k, \mathbf{w})\)
                   \For{\(j = 1\) \textbf{to} \(m\)}
                       \State \(\hat{V}^{\pi_k}_{j,\mathbf{w}} \gets V^{\pi_k}_{j,\mathbf{w}} + f^{k,j}_{\text{bagging}}(\mathbf{w})\)
                       \State \(\widetilde{V}^{\pi_k}_{j,\mathbf{w}} \gets \hat{V}^{\pi_k}_{j,\mathbf{w}} + \beta_{t'} \cdot \sigma_{k,j}(\mathbf{w})\)
                   \EndFor
                   \State Let \(\mathbf{V}_{\mathbf{w}} \gets \big(\widetilde{V}^{\pi_k}_{1,\mathbf{w}},\dots,\widetilde{V}^{\pi_k}_{m,\mathbf{w}}\big)\)
                   \State Compute \(HV \gets \texttt{HV}\Bigl(\texttt{Pareto}\bigl(\mathcal{L} \cup \{\mathbf{V}_{\mathbf{w}}\}\bigr)\Bigr)\), where \(\mathcal{L}\) is the set of existing objective vectors.
                   \State Add the pair \((\mathbf{w}, HV)\) to \(\mathcal{D}_k\)
               \EndFor
               \State Sort \(\mathcal{D}_k\) in descending order of \(HV\)
               \State Update working pool: \(\widetilde{\mathbf{W}}_k \gets\) top \(N\) weights from \(\mathcal{D}_k\)
           \EndFor
       \EndIf
   \EndFor
\end{algorithmic}
\end{algorithm}

\newpage

\section{Benchmark Problems}
This section provide detail objective return for each problems. Where $C$ in the following equations is live bonus.

\subsection{Swimmer-v2}
Observation and action space: $\mathcal{S} \in \mathbb{R}^{8},\mathcal{A} \in \mathbb{R}^{2}$. \\
The first objective is forward speed in x axis: 
\begin{equation}
    R_1 = v_x
\end{equation}
The second objective is energy efficiency:
\begin{equation}
    R_2 = 0.3-0.15\sum_i {a_i}^2, \quad a_i \in (-1,1)
\end{equation}

\subsection{HalfCheetah-v2}
Observation and action space: $\mathcal{S} \in \mathbb{R}^{17},\mathcal{A} \in \mathbb{R}^{6}$. \\
The first objective is forward speed in x axis: 
\begin{equation}
    R_1 = \min(v_x,4)+C
\end{equation}
The second objective is energy efficiency:
\begin{equation}
    R_2 = 4-\sum_i {a_i}^2+C, \quad a_i \in (-1,1)
\end{equation}
\begin{equation}
    C= 1
\end{equation}

\subsection{Walker2d-v2}
Observation and action space: $\mathcal{S} \in \mathbb{R}^{17},\mathcal{A} \in \mathbb{R}^{6}$. \\
The first objective is forward speed in x axis: 
\begin{equation}
    R_1 = v_x+C
\end{equation}
The second objective is energy efficiency:
\begin{equation}
    R_2 = 4-\sum_i {a_i}^2+C, \quad a_i \in (-1,1)
\end{equation}
\begin{equation}
    C= 1
\end{equation}

\subsection{Ant-v2}
Observation and action space: $\mathcal{S} \in \mathbb{R}^{27},\mathcal{A} \in \mathbb{R}^{8}$. \\
The first objective is forward speed in x axis: 
\begin{equation}
    R_1 = v_x+C
\end{equation}
The second objective is forward in y axis:
\begin{equation}
     R_2 = v_y+C
\end{equation}

\begin{equation}
    C= 1-0.5 \sum_ia_{i}^2, \quad a_i \in (-1,1)
\end{equation}

\subsection{Hopper-v2}
Observation and action space: $\mathcal{S} \in \mathbb{R}^{11},\mathcal{A} \in \mathbb{R}^{3}$. \\
The first objective is forward speed in x axis: 
\begin{equation}
    R_1 = 1.5v_x+C
\end{equation}
The second objective is jumping height:
\begin{equation}
    R_2 = 12(h-h_{init})+C
\end{equation}

\begin{equation}
    C= 1-2e^{-4} \sum_ia_{i}^2, \quad a_i \in (-1,1)
\end{equation}

\subsection{Hopper-v3}
Observation and action space: $\mathcal{S} \in \mathbb{R}^{11},\mathcal{A} \in \mathbb{R}^{3}$. \\
The first objective is forward speed in x axis: 
\begin{equation}
    R_1 = 1.5v_x+C
\end{equation}

The second objective is jumping height:
\begin{equation}
    R_2 = 12(h-h_{init})+C
\end{equation}

The third objective is energy efficiency:
\begin{equation}
    R_3 = 4-\sum_i a_{i}^2 +C
\end{equation}

\begin{equation}
    C= 1
\end{equation}

\section{PPO Hyperparameters}
\vspace{-0.75cm}
\begin{table}[!ht]
\centering
\caption{Hyper-parameter configuration of MOPPO algorithms.}
\label{tab:experimental_config}
\begin{tabular}{c|c} 
\hline
\textbf{Hyperparameters} & \textbf{Value} \\ \hline
Policy Number & 10 \\ \hline
Max Training Iterations & $2 \times 10^6$ \\ \hline
Number of Cells & 64 \\ \hline
Actor Learning Rate & $3 \times 10^{-4}$ \\ \hline
Critic Learning Rate & $3 \times 10^{-4}$ \\ \hline
Memory Size & 2500 \\ \hline
K Epochs & 10 \\ \hline
Gamma & 0.99 \\ \hline
Lambda & 0.95 \\ \hline
C1 Coefficient & 0.5 \\ \hline
C2 Coefficient & 0 \\ \hline
Epsilon Clip & 0.2 \\ \hline
Minibatch Size & 64 \\ \hline
\end{tabular}
\end{table}

\section{Convex Coverage Set Expansion}

In this section, we illustrate the expansion of the Convex Coverage Set (CCS) throughout the training process using our proposed UCB-MOPPO algorithm. The graphs are arranged sequentially from left to right and top to bottom, showing the progressive evolution of the CCS. Each graph depicts 100 sub-space vectors representing a two-objective optimization problem. The detailed comparison of baselines is provided in Table~\ref{tb:hv_eu_compare}, reporting the mean and standard deviation over three random seeds. As training progresses, a clear trend of CCS growth emerges, characterized by an increasing spread and coverage of vectors across the objective space. Notably, the distribution of vectors provides valuable insights: regions where vectors are more widely scattered indicate areas with fewer vectors dominated by others, reflecting an expansion toward a more optimal and comprehensive CCS. This separation demonstrates the growing diversity and coverage of the vectors over time, effectively showcasing the CCS’s ability to capture a wider range of trade-offs between objectives. Consequently, this illustrates the effectiveness of UCB-MOPPO in thoroughly exploring the objective space and improving the set of solutions throughout the training process.

% \begin{landscape}
\begin{table*}[!htbp]
\tiny
\caption{Evaluation of HV and EU metrics for continuous MORL tasks over three independent runs. The best results are highlighted in \textbf{bold}.
}
\centering

\setlength{\tabcolsep}{3pt} % Adjusts column gap globally
\renewcommand{\arraystretch}{1.8} % Adjusts row height (1.2-1.5 is usually good)

\label{tb:hv_eu_compare}
\begin{tabular}{cccccccccc} 
\textbf{Benchmark} & \textbf{Metric} & \textbf{UCB} & \textbf{Mean} & \textbf{Random} & \textbf{Fixed} & \textbf{PGMORL} & \textbf{PDMORL} & \textbf{CAPQL} & \textbf{GPI-LS} \\\hline
\multirow{2}{*}{Swimmer-V2} & HV ($10^4$)   & \textbf{5.60} $\pm$ \scriptsize{.18} & 4.72 $\pm$ \scriptsize{.19} & 4.56 $\pm$ \scriptsize{.23} & 4.45 $\pm$ \scriptsize{.11} & 1.67 $\pm$ \scriptsize{.09} & 1.77 $\pm$ \scriptsize{.02} & 2.40 $\pm$ \scriptsize{.03} & 4.75 $\pm$ \scriptsize{.02} \\
                  & EU ($10^2$)   & \textbf{2.18} $\pm$ \scriptsize{.48} & 2.16 $\pm$ \scriptsize{.25} & 2.08 $\pm$ \scriptsize{.22} & 2.11 $\pm$ \scriptsize{.12} & 1.23 $\pm$ \scriptsize{.42} & 1.24 $\pm$ \scriptsize{.49} & 1.79 $\pm$ \scriptsize{.45} & 2.14 $\pm$ \scriptsize{.41} \\\hline
\multirow{2}{*}{Halfcheetah-V2} & HV ($10^7$)   & 1.78 $\pm$ \scriptsize{.23}          & 1.60 $\pm$ \scriptsize{.07} & 1.21 $\pm$ \scriptsize{.09} & 1.12 $\pm$ \scriptsize{.05} & 0.58 $\pm$ \scriptsize{.01} & 0.62 $\pm$ \scriptsize{.02} & \textbf{2.20} $\pm$ \scriptsize{.08} & 2.16 $\pm$ \scriptsize{.02} \\
                  & EU ($10^3$)   & 3.88 $\pm$ \scriptsize{.22}          & 3.58 $\pm$ \scriptsize{.25} & 3.55 $\pm$ \scriptsize{.30} & 3.44 $\pm$ \scriptsize{.36} & 2.30 $\pm$ \scriptsize{.44} & 2.42 $\pm$ \scriptsize{.31} & \textbf{4.47} $\pm$ \scriptsize{.03} & 4.30 $\pm$ \scriptsize{.08} \\\hline
\multirow{2}{*}{Walker2d-V2} & HV ($10^7$)   & \textbf{1.40} $\pm$ \scriptsize{.03} & 1.29 $\pm$ \scriptsize{.13} & 1.15 $\pm$ \scriptsize{.06} & 1.13 $\pm$ \scriptsize{.07} & 0.44 $\pm$ \scriptsize{.02} & 0.56 $\pm$ \scriptsize{.04} & 0.18 $\pm$ \scriptsize{.05} & 1.22 $\pm$ \scriptsize{.04} \\
                  & EU ($10^3$)   & \textbf{3.54} $\pm$ \scriptsize{.42} & 3.42 $\pm$ \scriptsize{.29} & 3.34 $\pm$ \scriptsize{.29} & 3.30 $\pm$ \scriptsize{.33} & 1.98 $\pm$ \scriptsize{.21} & 2.24 $\pm$ \scriptsize{.34} & 1.50 $\pm$ \scriptsize{.12} & 3.40 $\pm$ \scriptsize{.30} \\\hline
\multirow{2}{*}{Ant-V2} & HV ($10^7$)   & \textbf{1.07} $\pm$ \scriptsize{.09} & 0.92 $\pm$ \scriptsize{.04} & 0.65 $\pm$ \scriptsize{.01} & 0.60 $\pm$ \scriptsize{.01} & 0.61 $\pm$ \scriptsize{.10} & 0.66 $\pm$ \scriptsize{.02} & 0.45 $\pm$ \scriptsize{.04} & 0.81 $\pm$ \scriptsize{.01} \\
                  & EU ($10^3$)   & \textbf{2.96} $\pm$ \scriptsize{.43} & 2.61 $\pm$ \scriptsize{.17} & 2.35 $\pm$ \scriptsize{.24} & 2.21 $\pm$ \scriptsize{.39} & 2.28 $\pm$ \scriptsize{.35} & 2.41 $\pm$ \scriptsize{.20} & 2.03 $\pm$ \scriptsize{.42} & 2.84 $\pm$ \scriptsize{.26} \\\hline
\multirow{2}{*}{Hopper-V2} & HV ($10^7$)   & \textbf{0.84} $\pm$ \scriptsize{.05} & 0.81 $\pm$ \scriptsize{.04} & 0.79 $\pm$ \scriptsize{.01} & 0.76 $\pm$ \scriptsize{.02} & 0.23 $\pm$ \scriptsize{.02} & 0.25 $\pm$ \scriptsize{.02} & 0.25 $\pm$ \scriptsize{.07} & 0.80 $\pm$ \scriptsize{.02} \\
                  & EU ($10^3$)   & 2.77 $\pm$ \scriptsize{.28} & \textbf{2.84} $\pm$ \scriptsize{.30} & 2.77 $\pm$ \scriptsize{.18} & 2.71 $\pm$ \scriptsize{.26} & 1.48 $\pm$ \scriptsize{.47} & 1.51 $\pm$ \scriptsize{.18} & 1.46 $\pm$ \scriptsize{.11} & 2.75 $\pm$ \scriptsize{.07} \\\hline
\multirow{2}{*}{Hopper-V3} & HV ($10^{10}$)& \textbf{3.62} $\pm$ \scriptsize{.01} & 2.83 $\pm$ \scriptsize{.22} & 2.90 $\pm$ \scriptsize{.11} & 2.64 $\pm$ \scriptsize{.18} & 0.63 $\pm$ \scriptsize{.07} & 0.11 $\pm$ \scriptsize{.02} & 0.31 $\pm$ \scriptsize{.06} & 0.65 $\pm$ \scriptsize{.03} \\
                  & EU ($10^3$)   & 3.20 $\pm$ \scriptsize{.02} & 3.19 $\pm$ \scriptsize{.04}  & \textbf{3.28} $\pm$ \scriptsize{.20}  & 2.90 $\pm$ \scriptsize{.01}  & 1.70 $\pm$ \scriptsize{.04}  & 1.74 $\pm$ \scriptsize{.09}  & 1.52  $\pm$ \scriptsize{.29}  & 2.14 $\pm$ \scriptsize{.42} \\\hline
\end{tabular}
\end{table*}
% \end{landscape}

% ####################################################################################
% ####################################################################################
\newpage

\subsection{Swimmer-V2}

\begin{figure}[H]
\centering
\vspace{-0.5cm}
% First Row
\begin{subfigure}{0.24\textwidth}
  \centering
  \includegraphics[width=\linewidth]{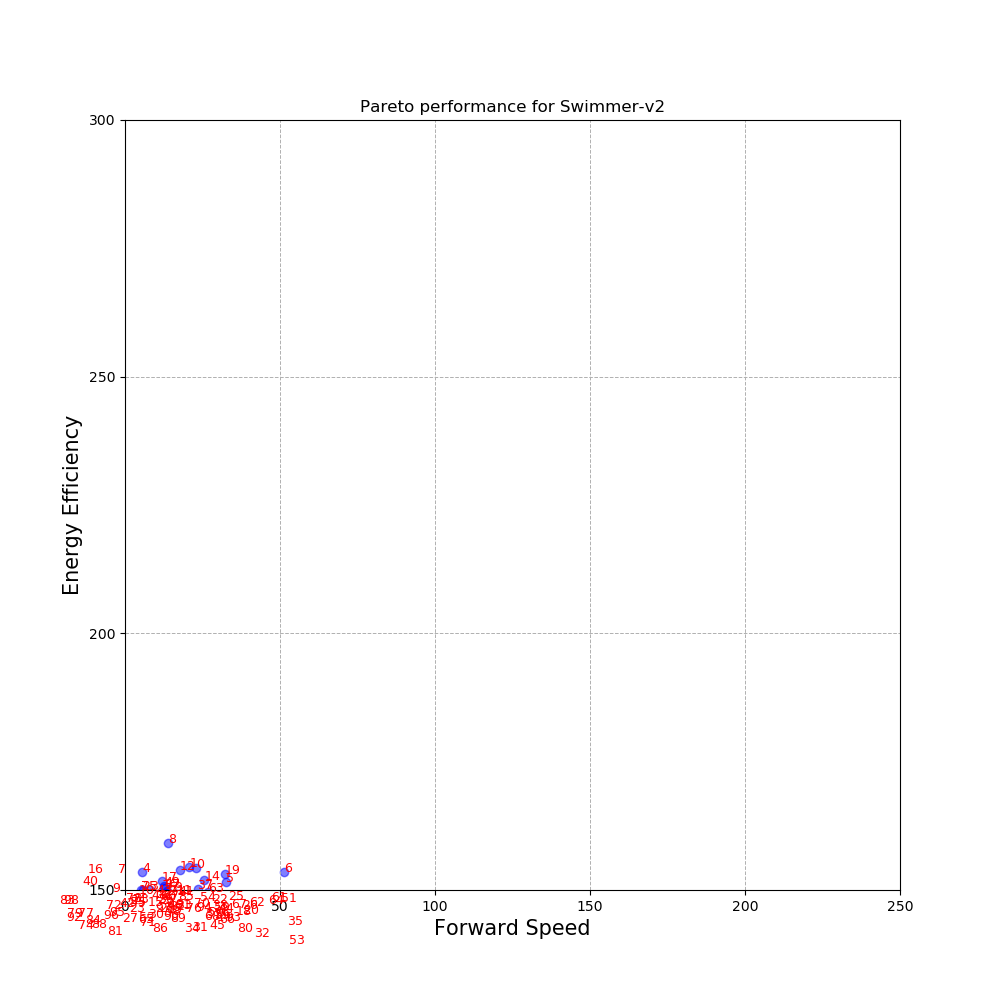}
  \label{fig:sub1}
\end{subfigure}
\begin{subfigure}{0.24\textwidth}
  \centering
  \includegraphics[width=\linewidth]{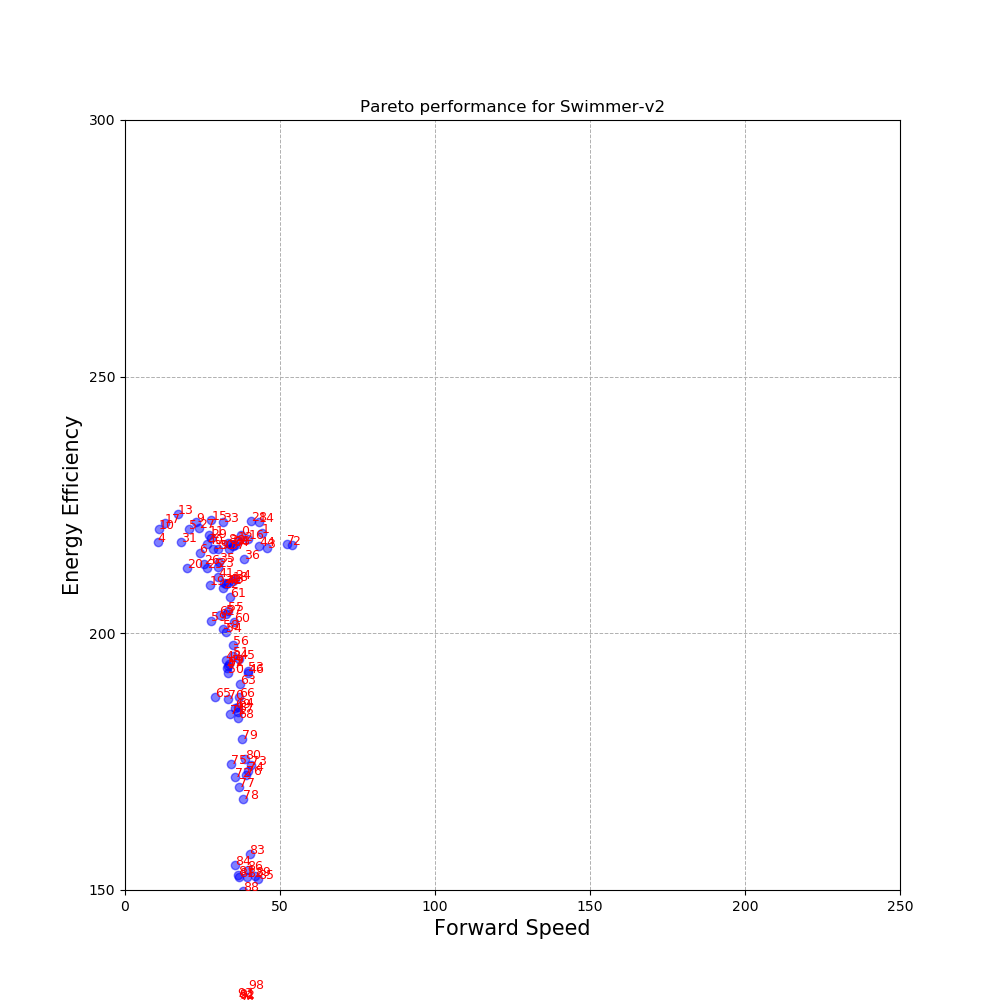}
  \label{fig:sub2}
\end{subfigure}
\begin{subfigure}{0.24\textwidth}
  \centering
  \includegraphics[width=\linewidth]{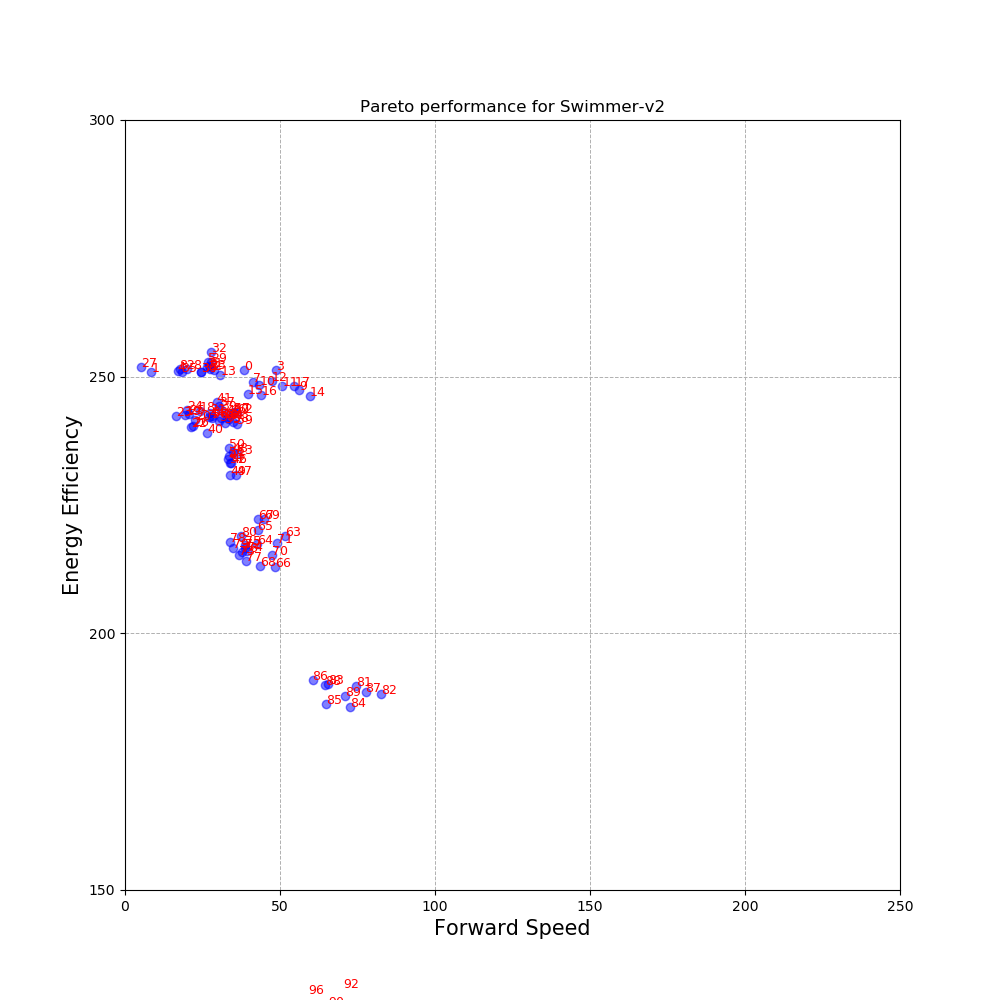}
  \label{fig:sub3}
\end{subfigure}
\begin{subfigure}{0.24\textwidth}
  \centering
  \includegraphics[width=\linewidth]{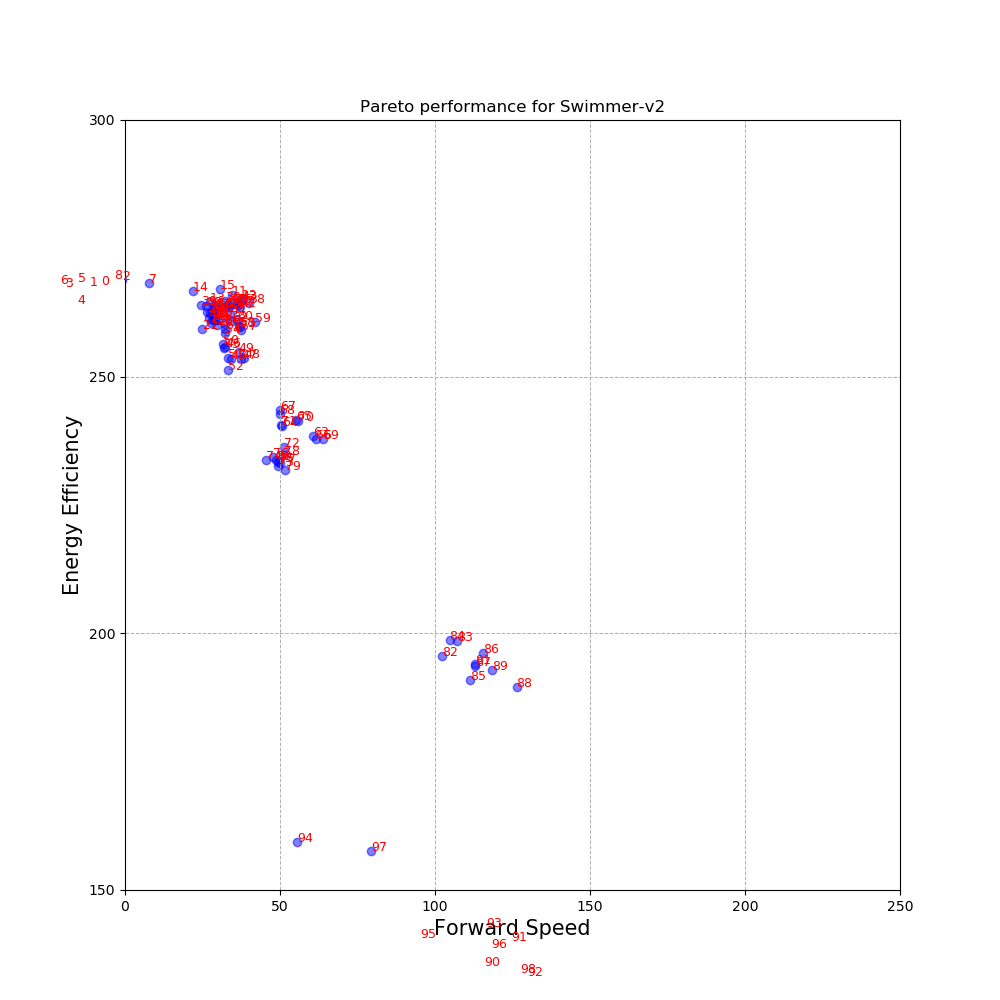}
  \label{fig:sub4}
\end{subfigure}

% Second Row
\begin{subfigure}{0.24\textwidth}
  \centering
  \includegraphics[width=\linewidth]{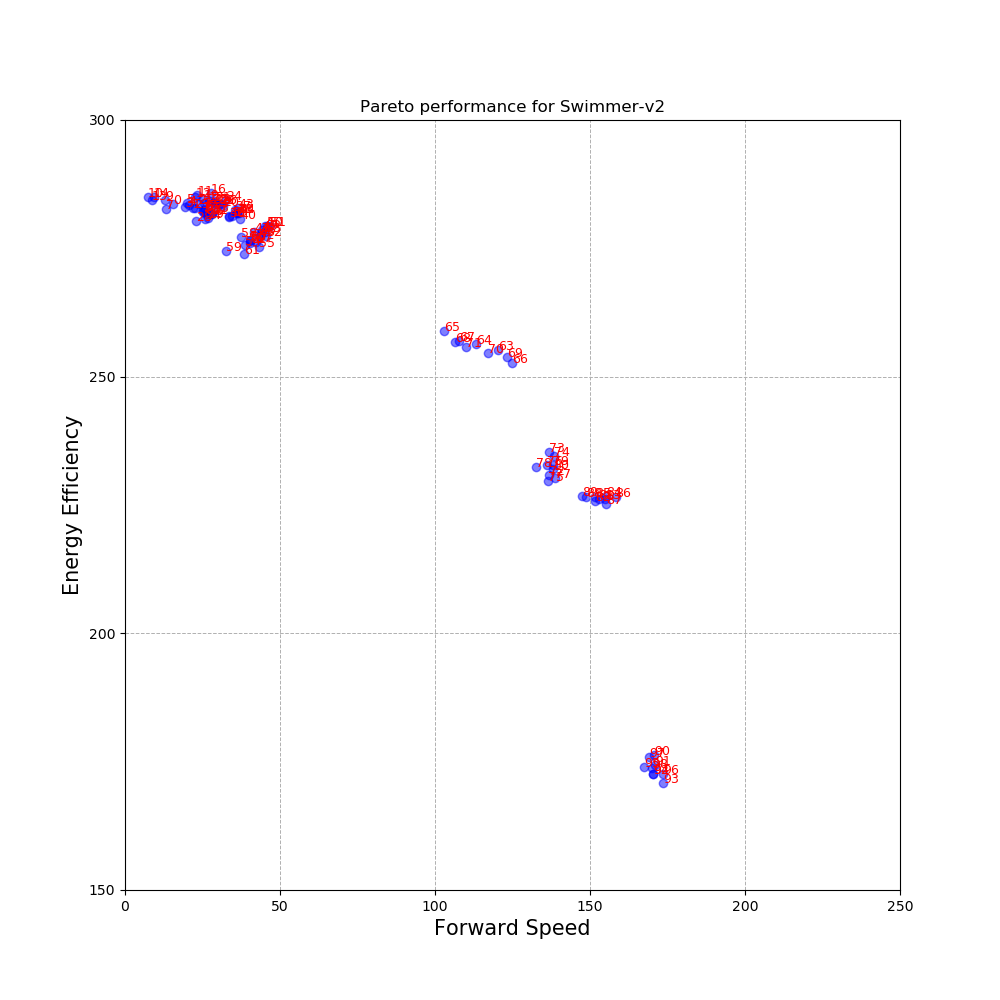}
  \label{fig:sub5}
\end{subfigure}
\begin{subfigure}{0.24\textwidth}
  \centering
  \includegraphics[width=\linewidth]{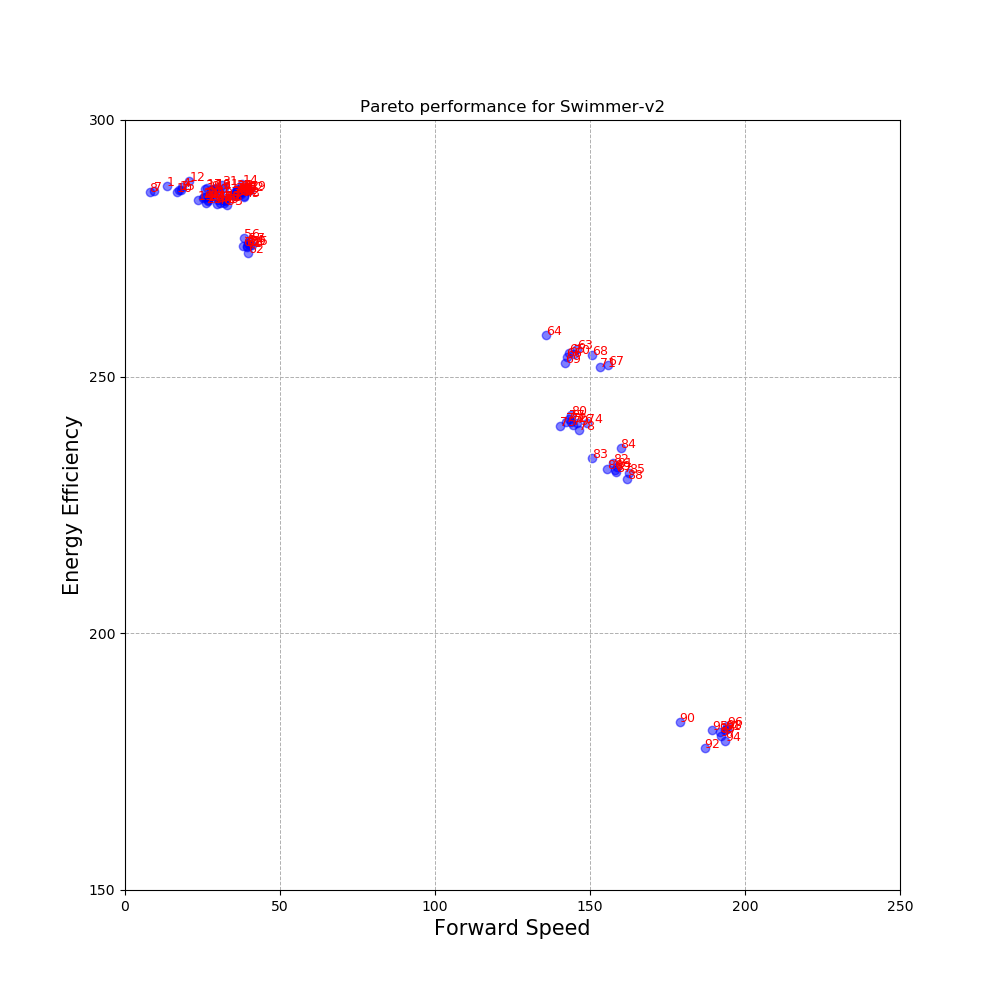}
  \label{fig:sub6}
\end{subfigure}
\begin{subfigure}{0.24\textwidth}
  \centering
  \includegraphics[width=\linewidth]{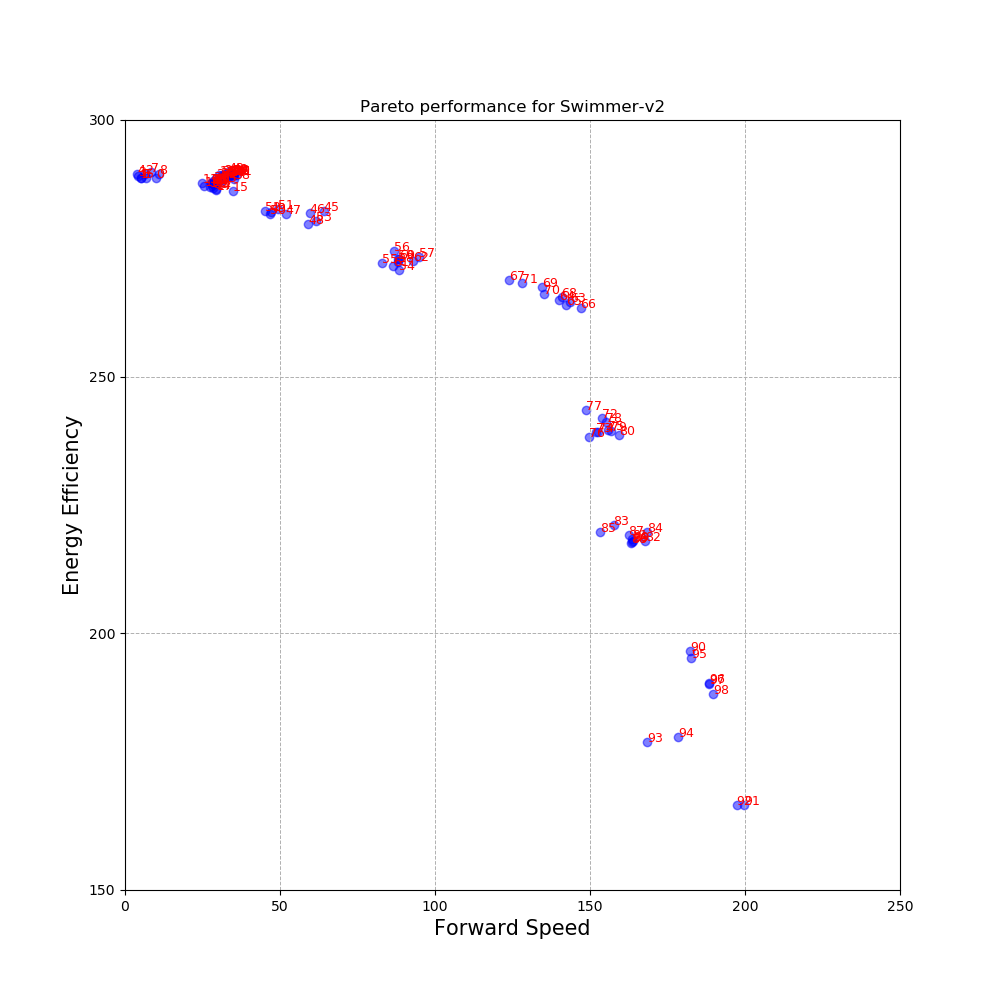}
  \label{fig:sub7}
\end{subfigure}
\begin{subfigure}{0.24\textwidth}
  \centering
  \includegraphics[width=\linewidth]{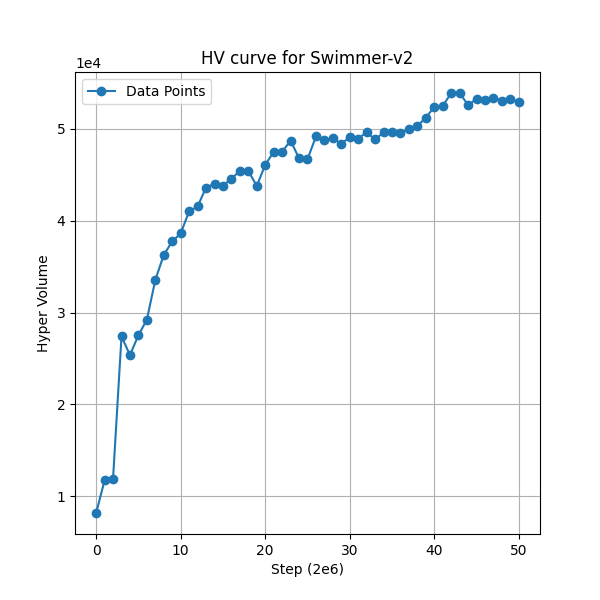}
  \label{fig:sub8}
\end{subfigure}

\caption{CCS expansion in Swimmer-V2 with one seed. The last graph shows the hypervolume growth.}
\label{fig:swimmer_ccs}

\end{figure}

% ####################################################################################
% ####################################################################################

\newpage
\vspace{2cm}
\subsection{HalfCheetah-v2}
\begin{figure}[H]
\centering
\vspace{-0.5cm}
% First Row
\begin{subfigure}{0.24\textwidth}
  \centering
  \includegraphics[width=\linewidth]{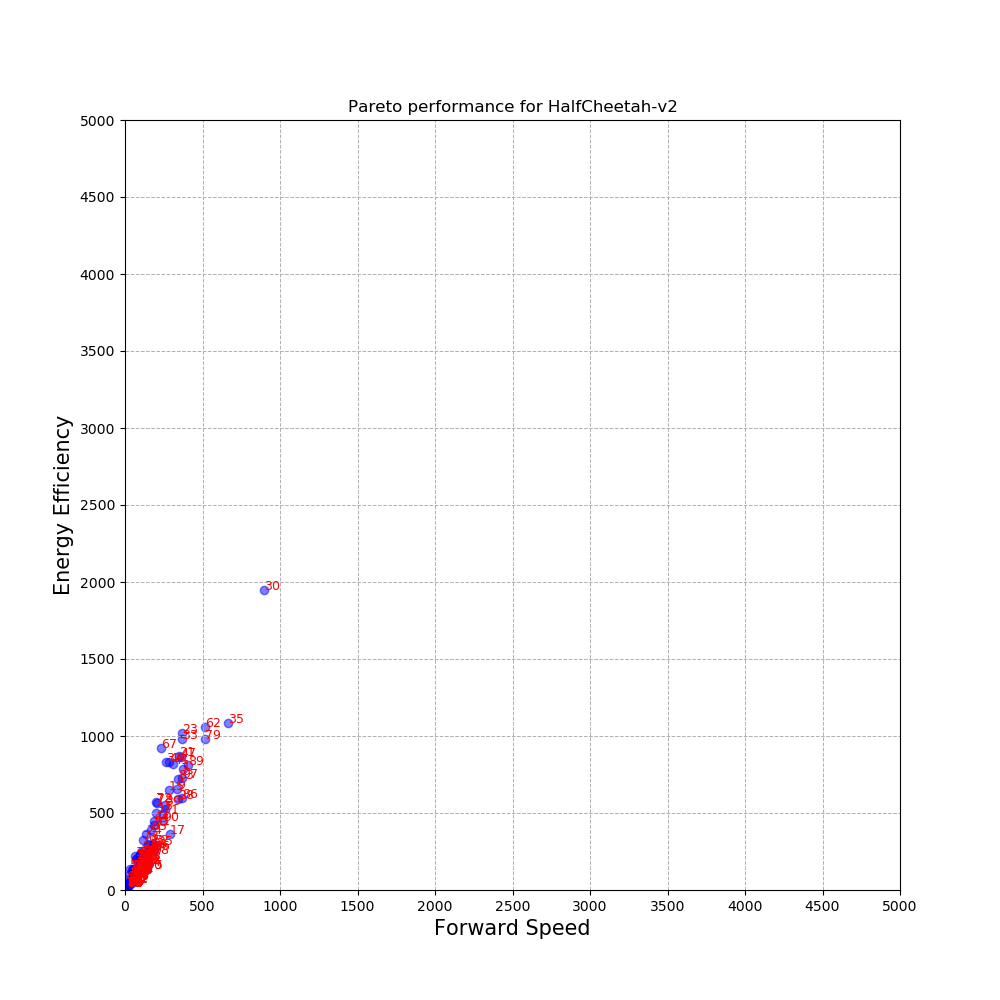}
  \label{fig:sub1}
\end{subfigure}
\begin{subfigure}{0.24\textwidth}
  \centering
  \includegraphics[width=\linewidth]{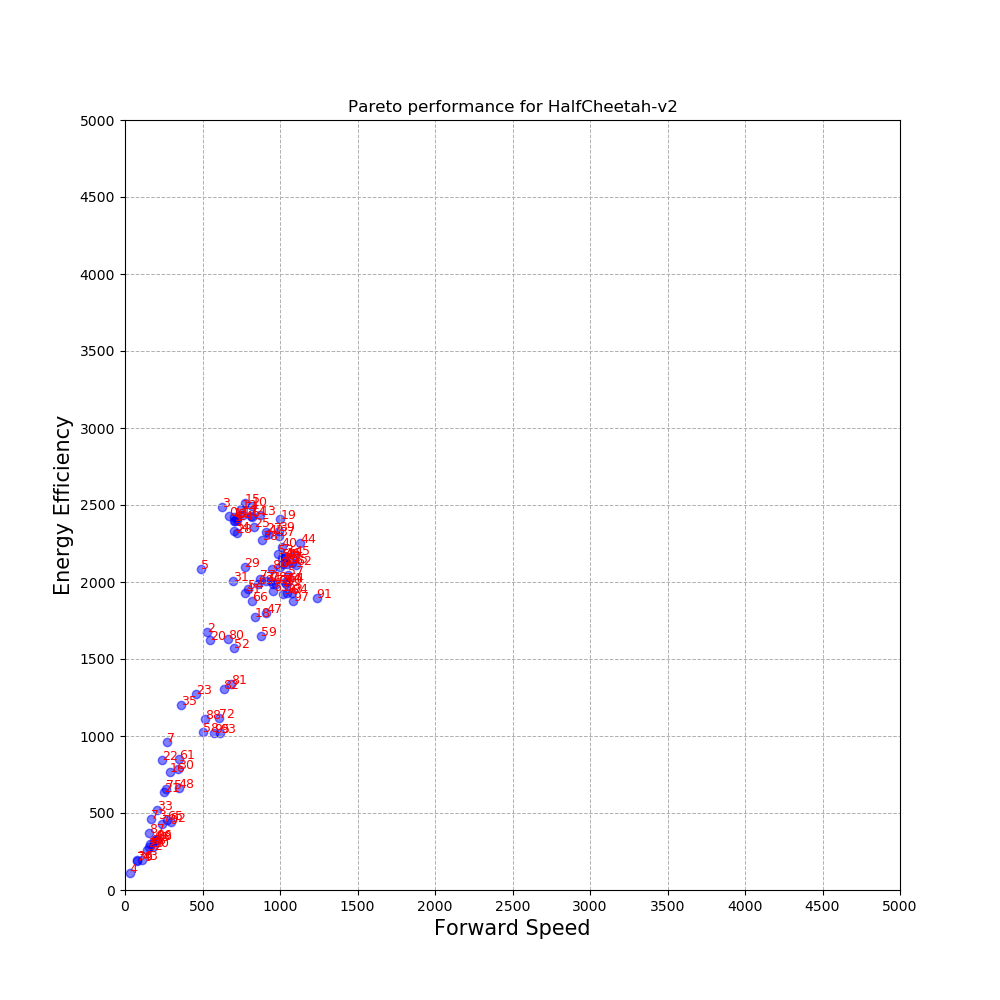}
  \label{fig:sub2}
\end{subfigure}
\begin{subfigure}{0.24\textwidth}
  \centering
  \includegraphics[width=\linewidth]{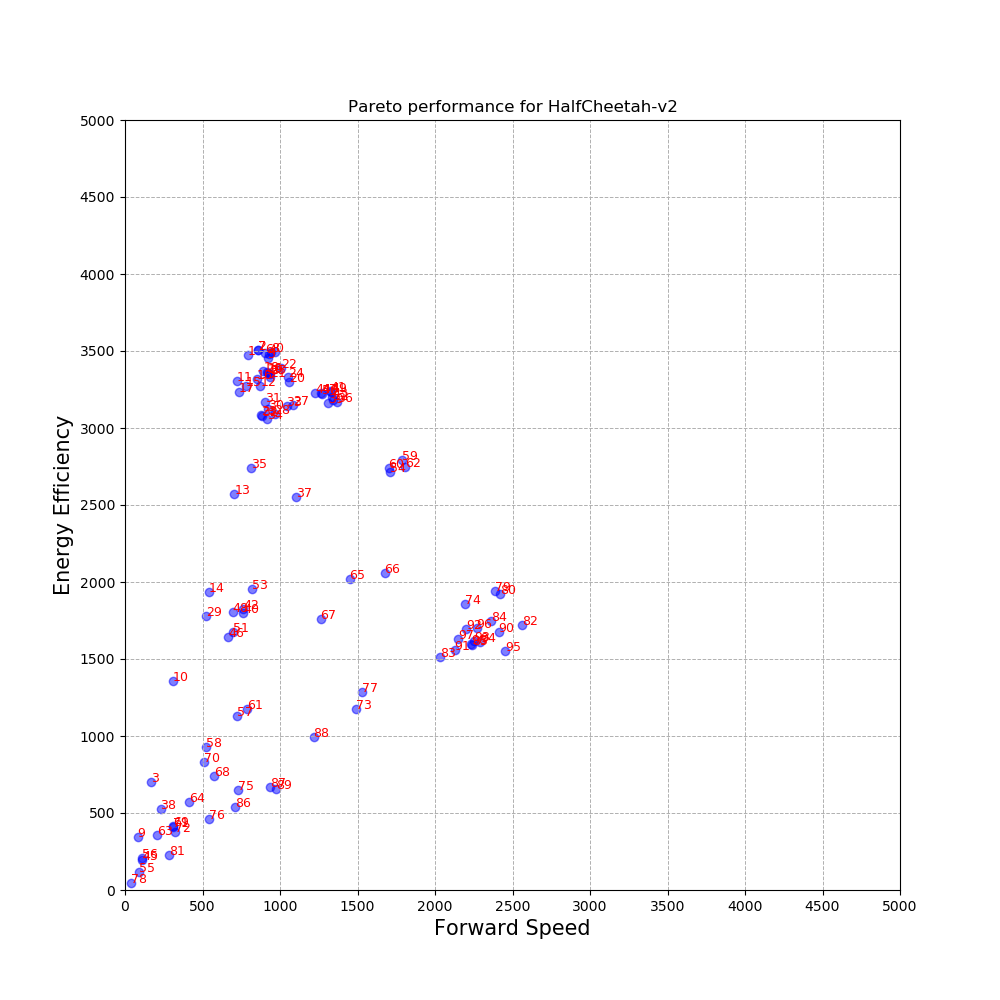}
  \label{fig:sub3}
\end{subfigure}
\begin{subfigure}{0.24\textwidth}
  \centering
  \includegraphics[width=\linewidth]{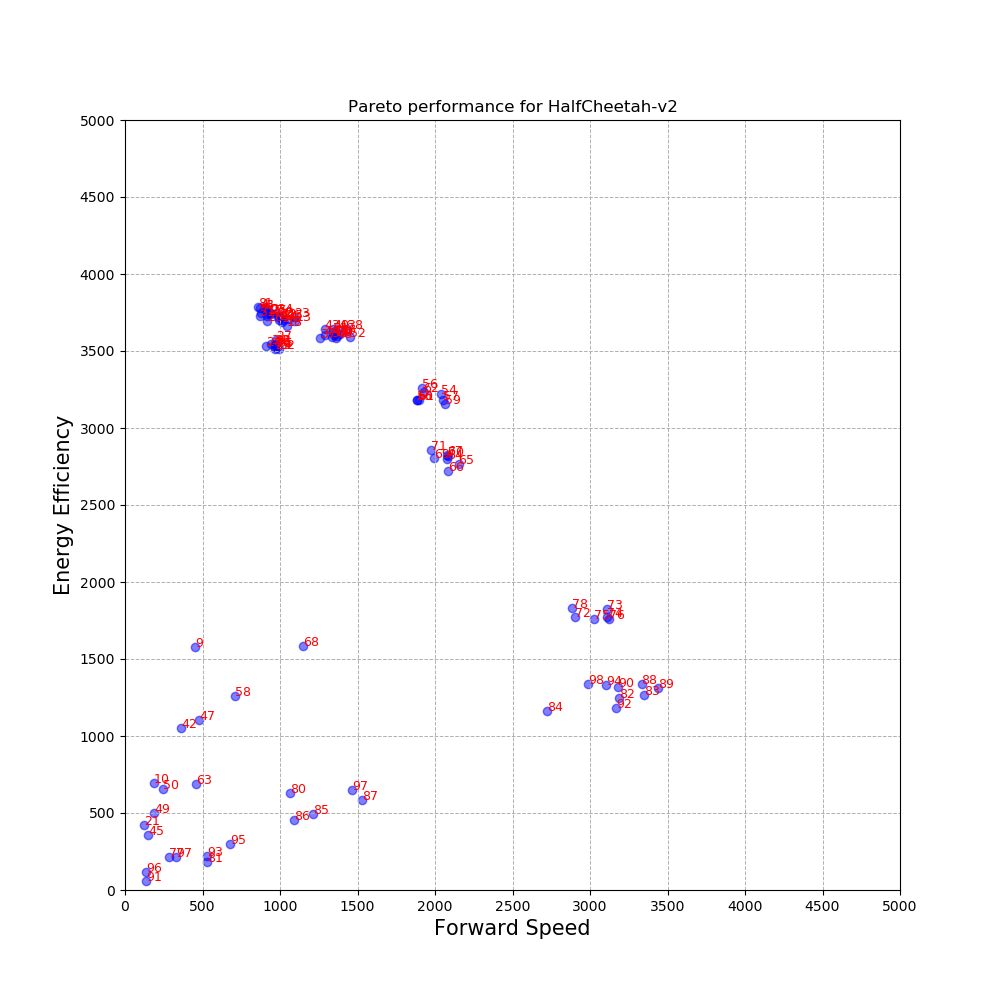}
  \label{fig:sub4}
\end{subfigure}

% Second Row
\begin{subfigure}{0.24\textwidth}
  \centering
  \includegraphics[width=\linewidth]{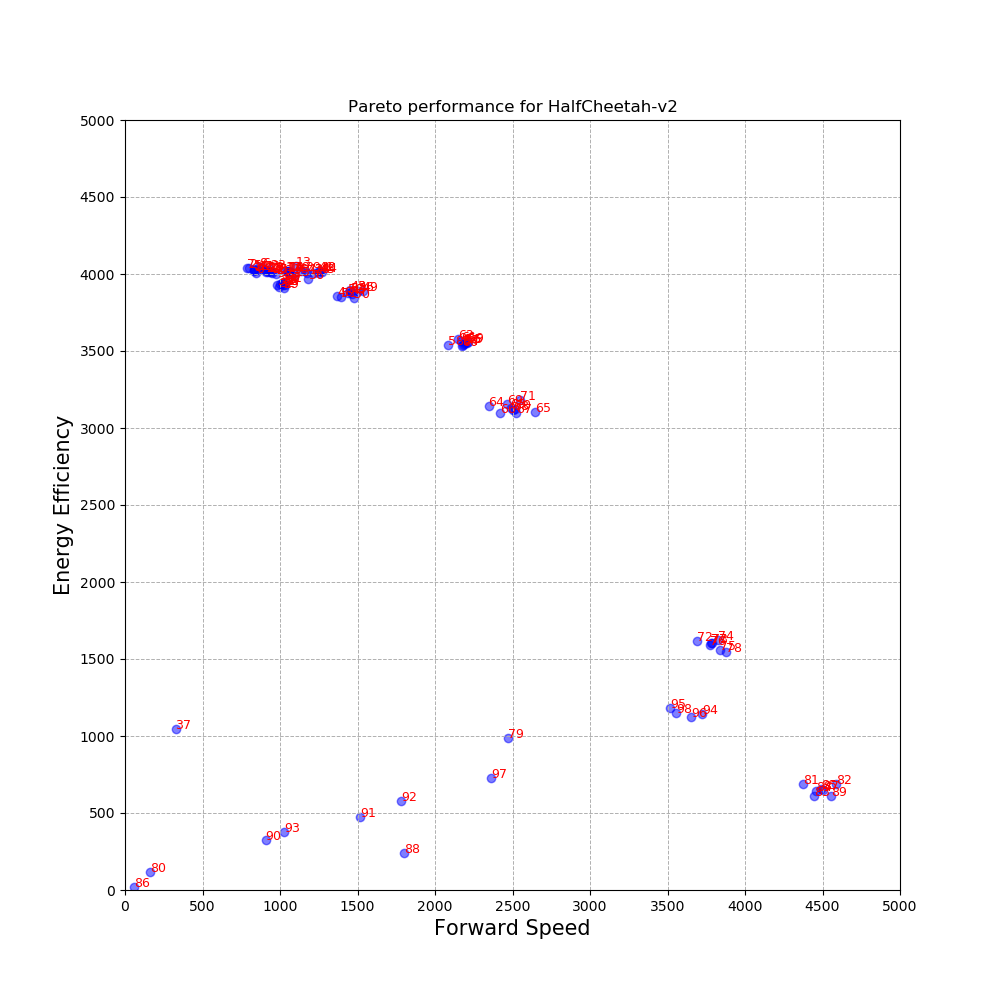}
  \label{fig:sub5}
\end{subfigure}
\begin{subfigure}{0.24\textwidth}
  \centering
  \includegraphics[width=\linewidth]{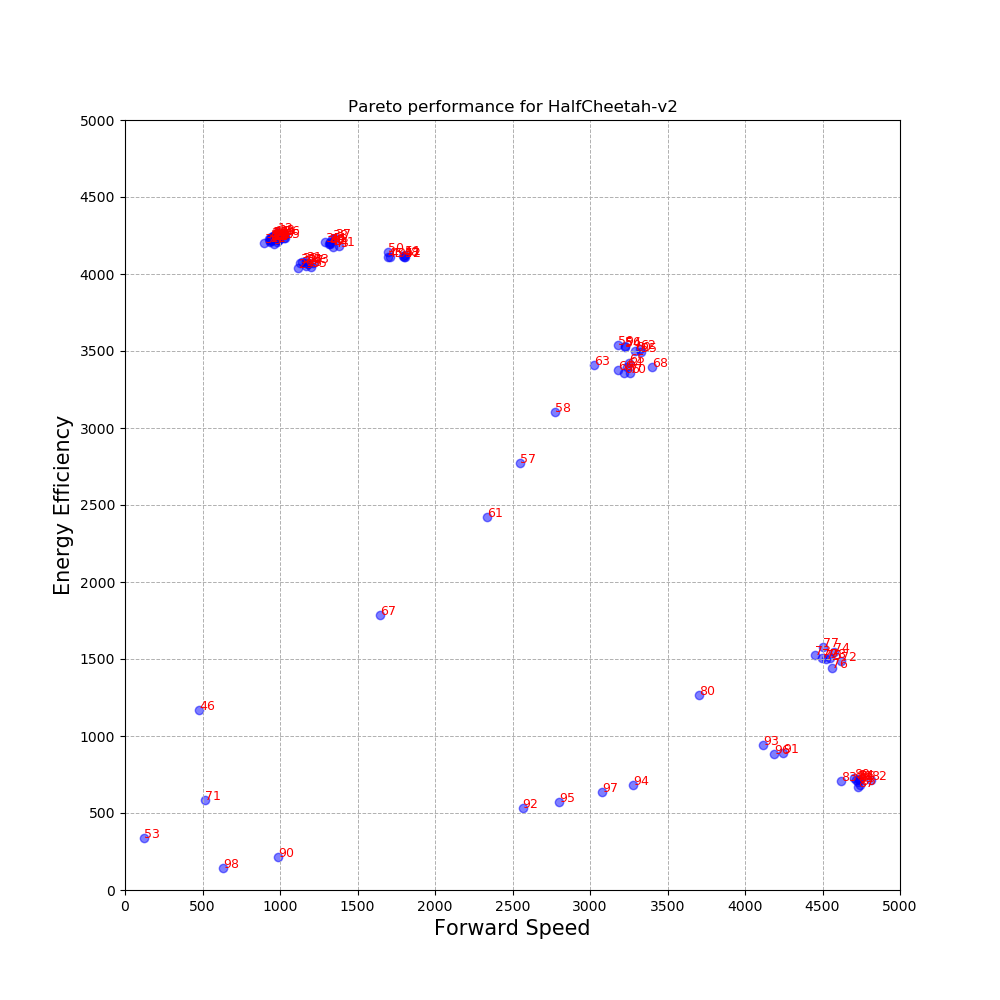}
  \label{fig:sub6}
\end{subfigure}
\begin{subfigure}{0.24\textwidth}
  \centering
  \includegraphics[width=\linewidth]{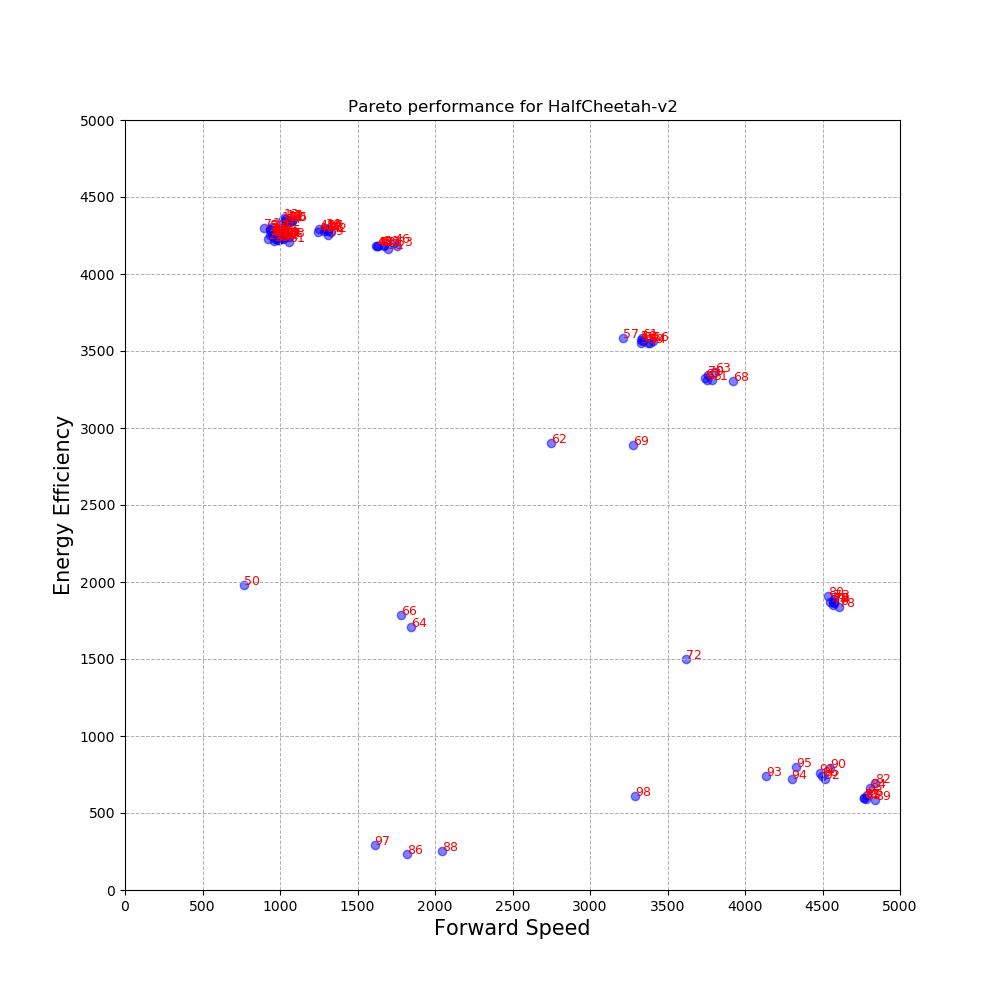}
  \label{fig:sub7}
\end{subfigure}
\begin{subfigure}{0.24\textwidth}
  \centering
  \includegraphics[width=\linewidth]{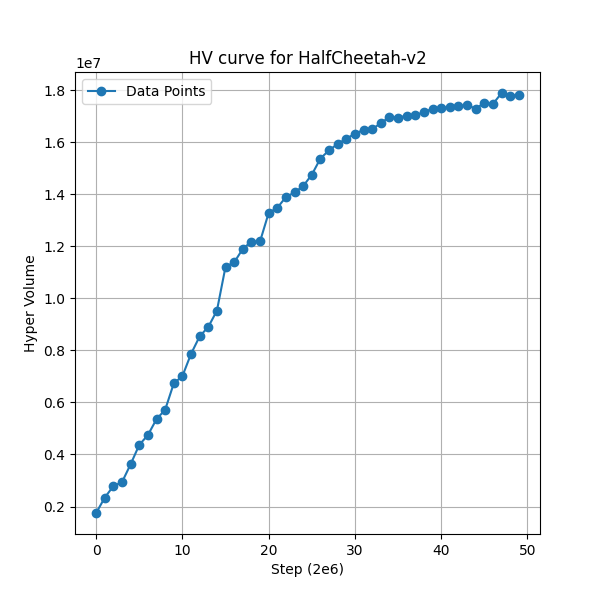}
  \label{fig:sub8}
\end{subfigure}

\caption{CCS expansion in HalfCheetah-V2 with one seed. The last graph shows the hypervolume growth.}
\label{fig:halfcheetah_ccs}

\end{figure}

% ####################################################################################
% ####################################################################################
\newpage
\subsection{Walker2D-V2}
\begin{figure}[H]
\centering
\vspace{-0.5cm}
% First Row
\begin{subfigure}{0.24\textwidth}
  \centering
  \includegraphics[width=\linewidth]{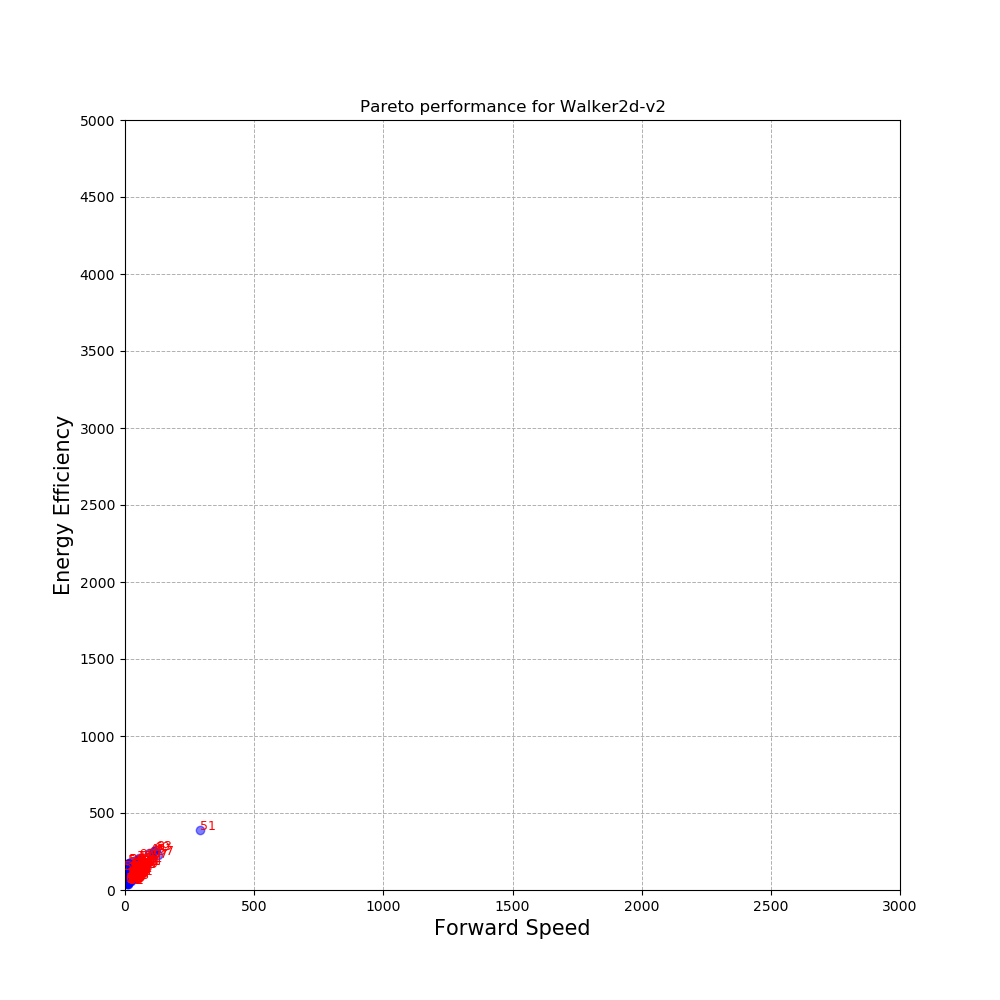}
  \label{fig:sub1}
\end{subfigure}
\begin{subfigure}{0.24\textwidth}
  \centering
  \includegraphics[width=\linewidth]{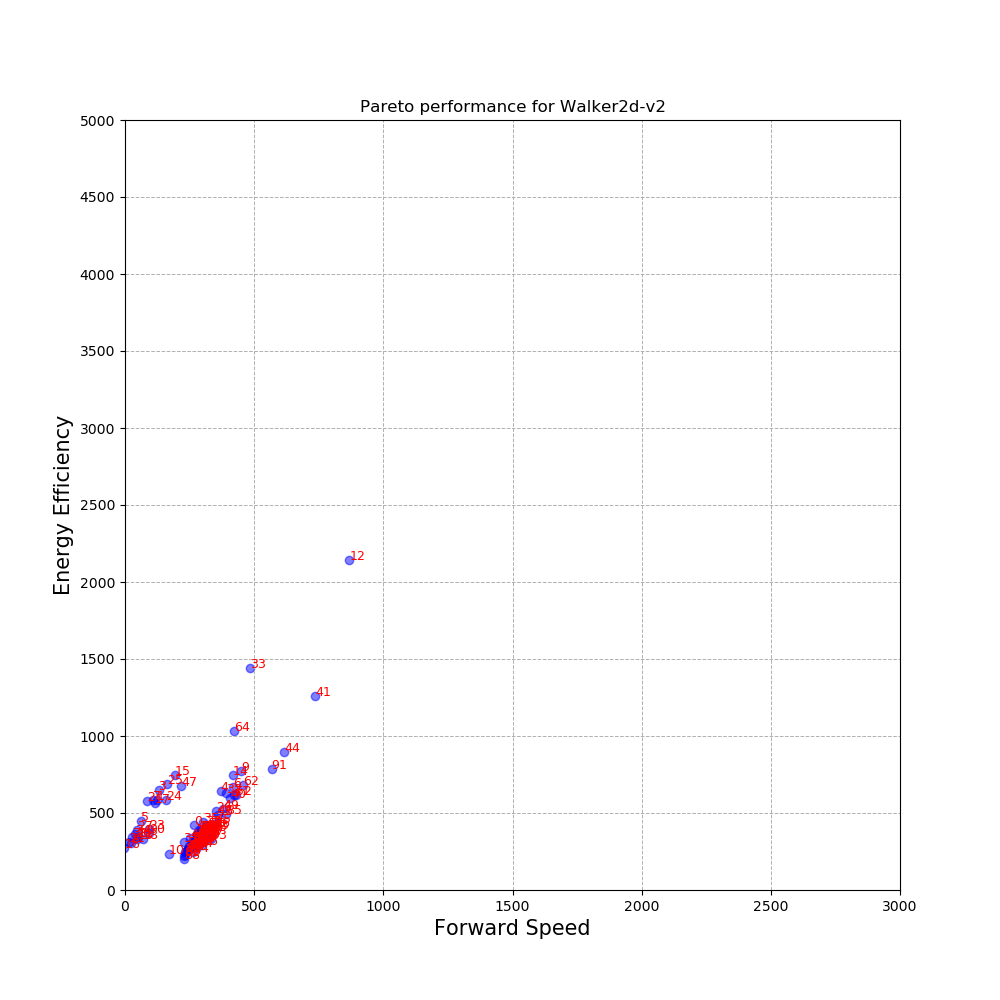}
  \label{fig:sub2}
\end{subfigure}
\begin{subfigure}{0.24\textwidth}
  \centering
  \includegraphics[width=\linewidth]{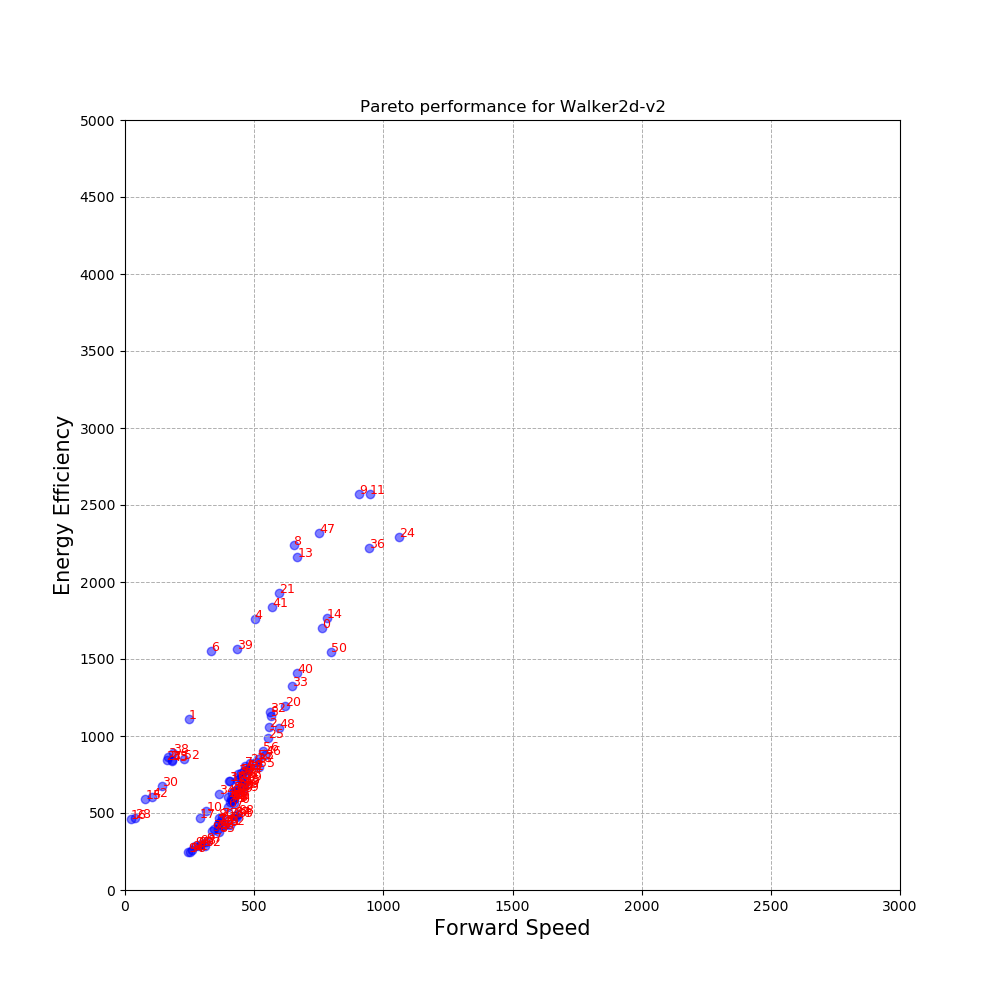}
  \label{fig:sub3}
\end{subfigure}
\begin{subfigure}{0.24\textwidth}
  \centering
  \includegraphics[width=\linewidth]{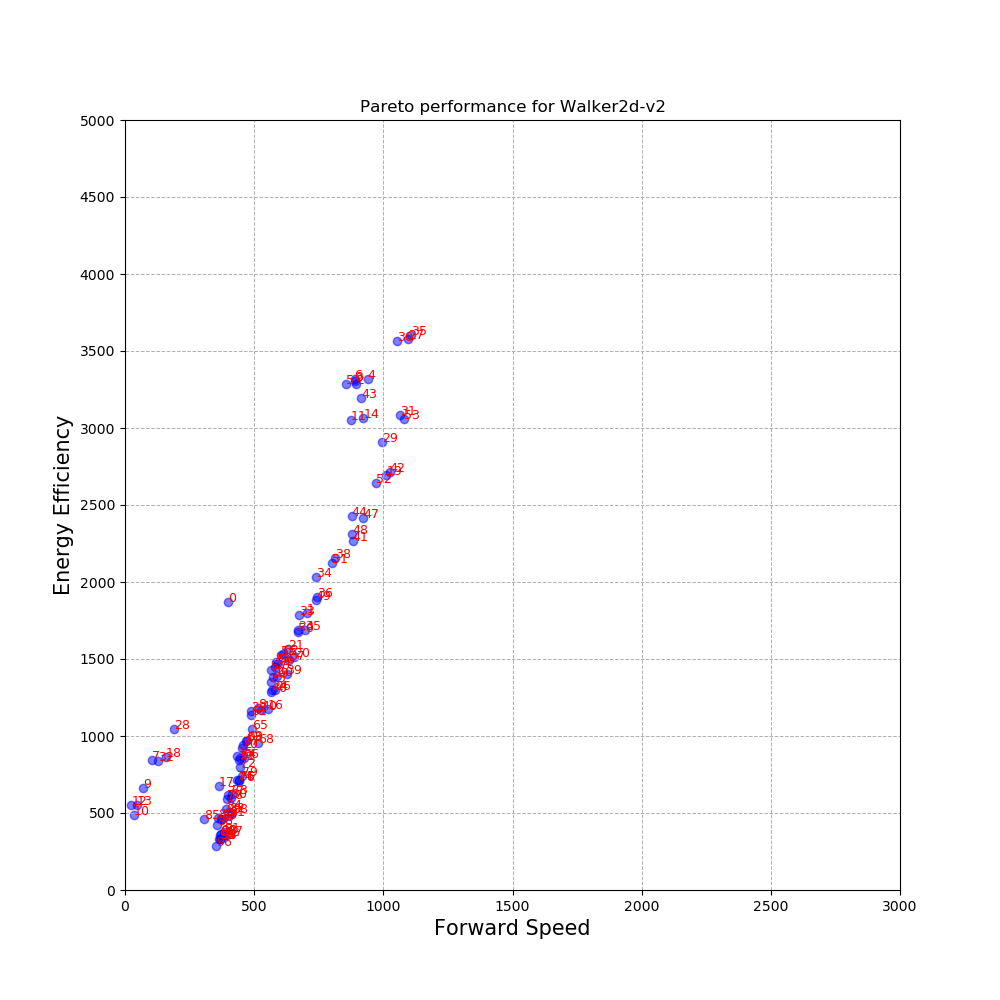}
  \label{fig:sub4}
\end{subfigure}

% Second Row
\begin{subfigure}{0.24\textwidth}
  \centering
  \includegraphics[width=\linewidth]{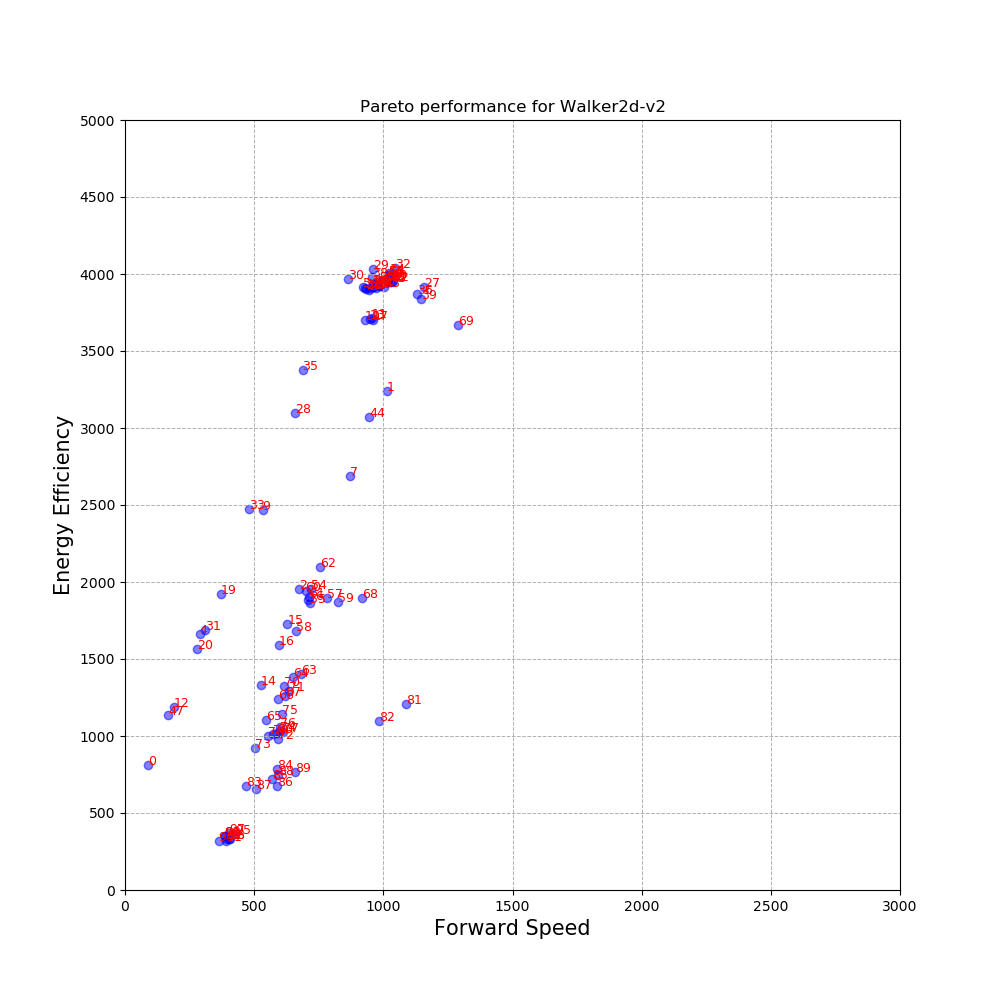}
  \label{fig:sub5}
\end{subfigure}
\begin{subfigure}{0.24\textwidth}
  \centering
  \includegraphics[width=\linewidth]{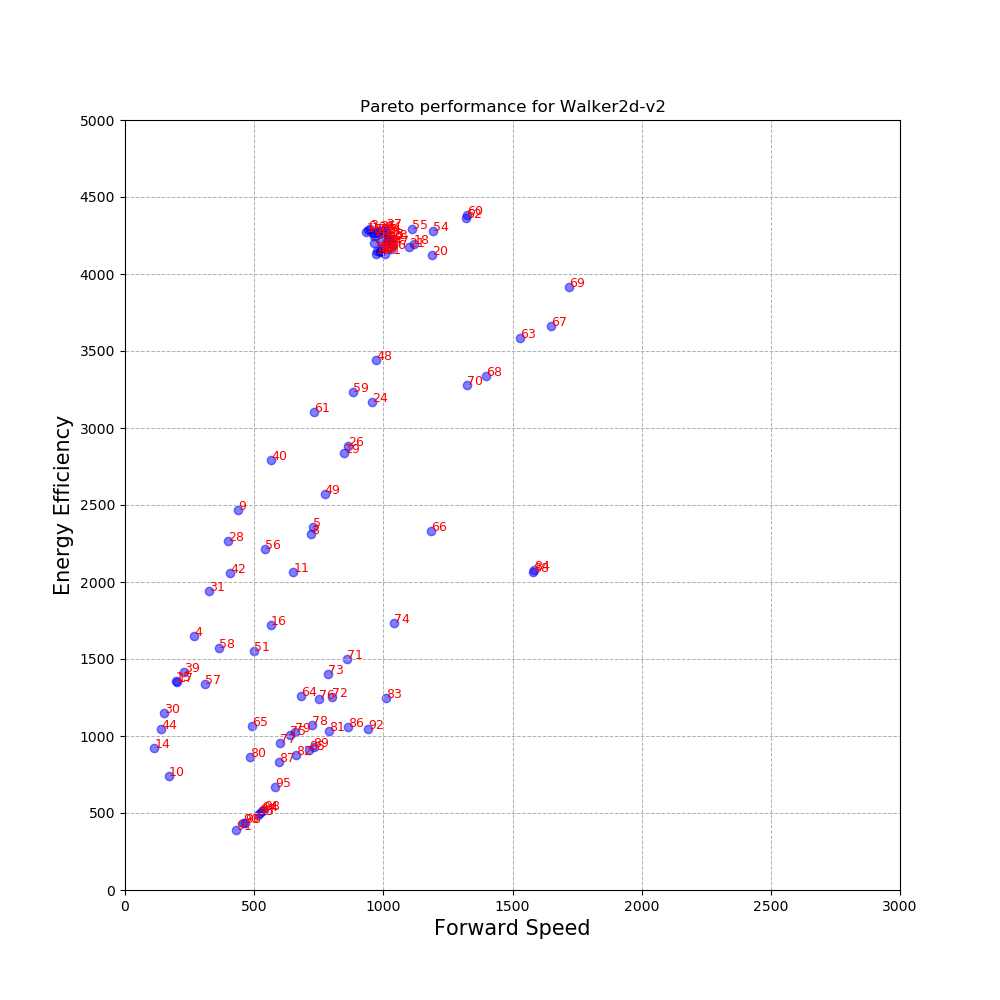}
  \label{fig:sub6}
\end{subfigure}
\begin{subfigure}{0.24\textwidth}
  \centering
  \includegraphics[width=\linewidth]{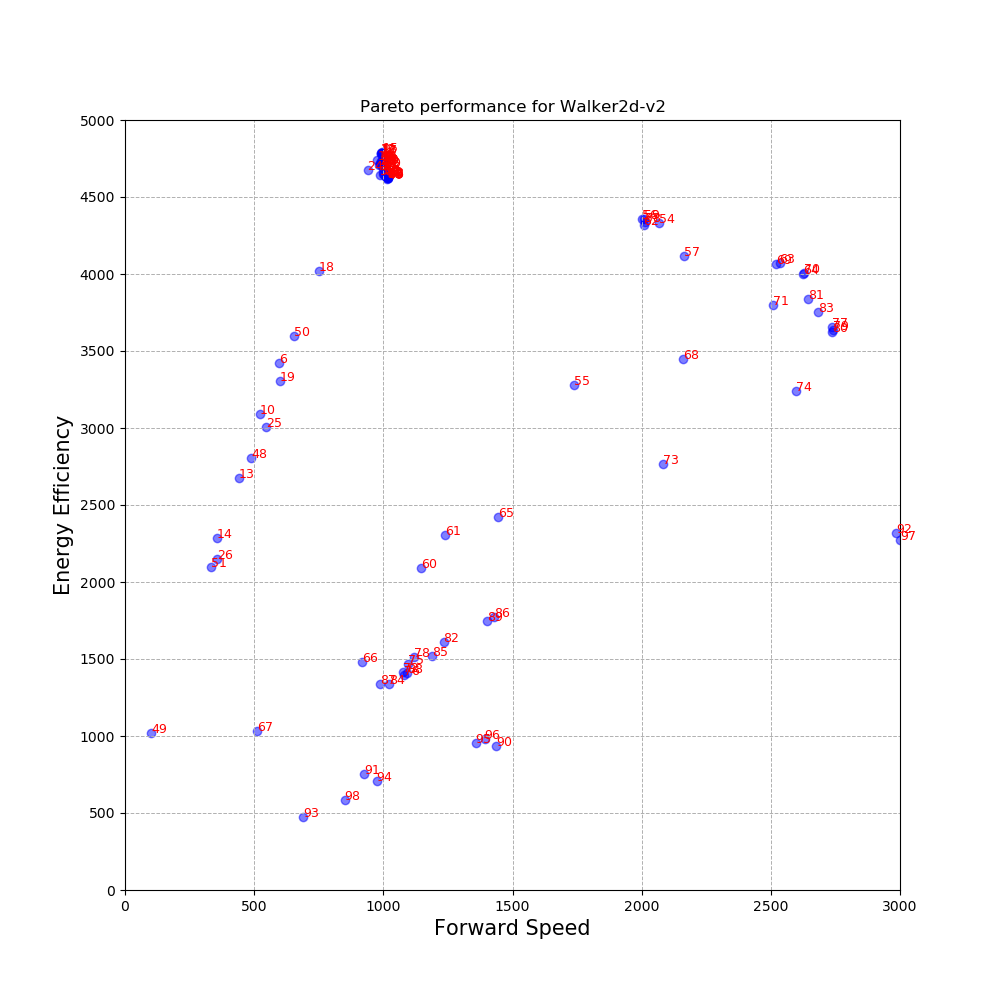}
  \label{fig:sub7}
\end{subfigure}
\begin{subfigure}{0.24\textwidth}
  \centering
  \includegraphics[width=\linewidth]{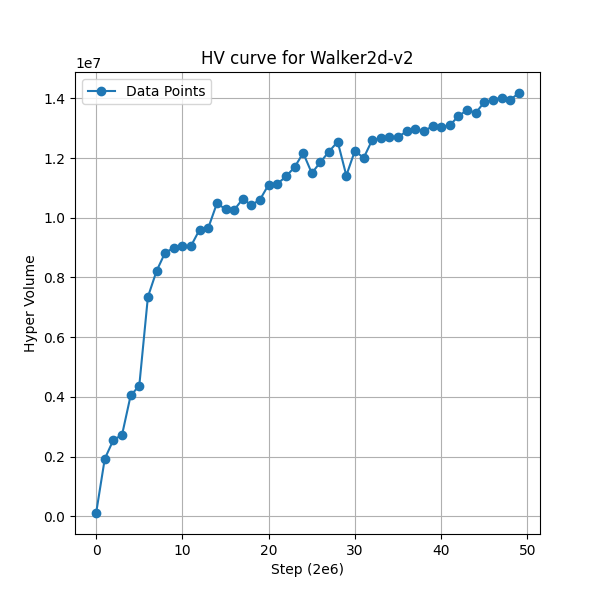}
  \label{fig:sub8}
\end{subfigure}

\caption{CCS expansion in Walker2D-V2 with one seed. The last graph shows the hypervolume growth.}
\label{fig:walker2d_ccs}

\end{figure}

% ####################################################################################
% ####################################################################################
\newpage
\subsection{Ant-V2}
\begin{figure}[H]
\centering
\vspace{-0.5cm}
% First Row
\begin{subfigure}{0.24\textwidth}
  \centering
  \includegraphics[width=\linewidth]{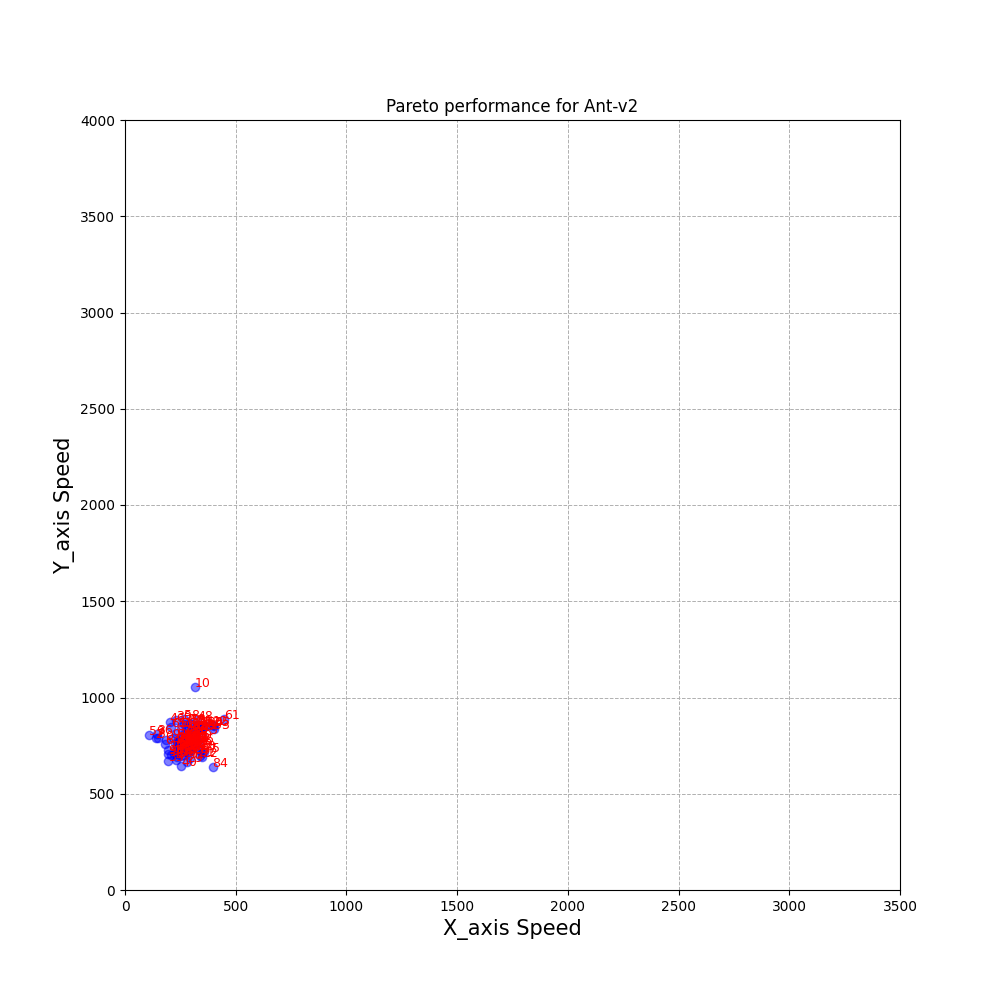}
  \label{fig:sub1}
\end{subfigure}
\begin{subfigure}{0.24\textwidth}
  \centering
  \includegraphics[width=\linewidth]{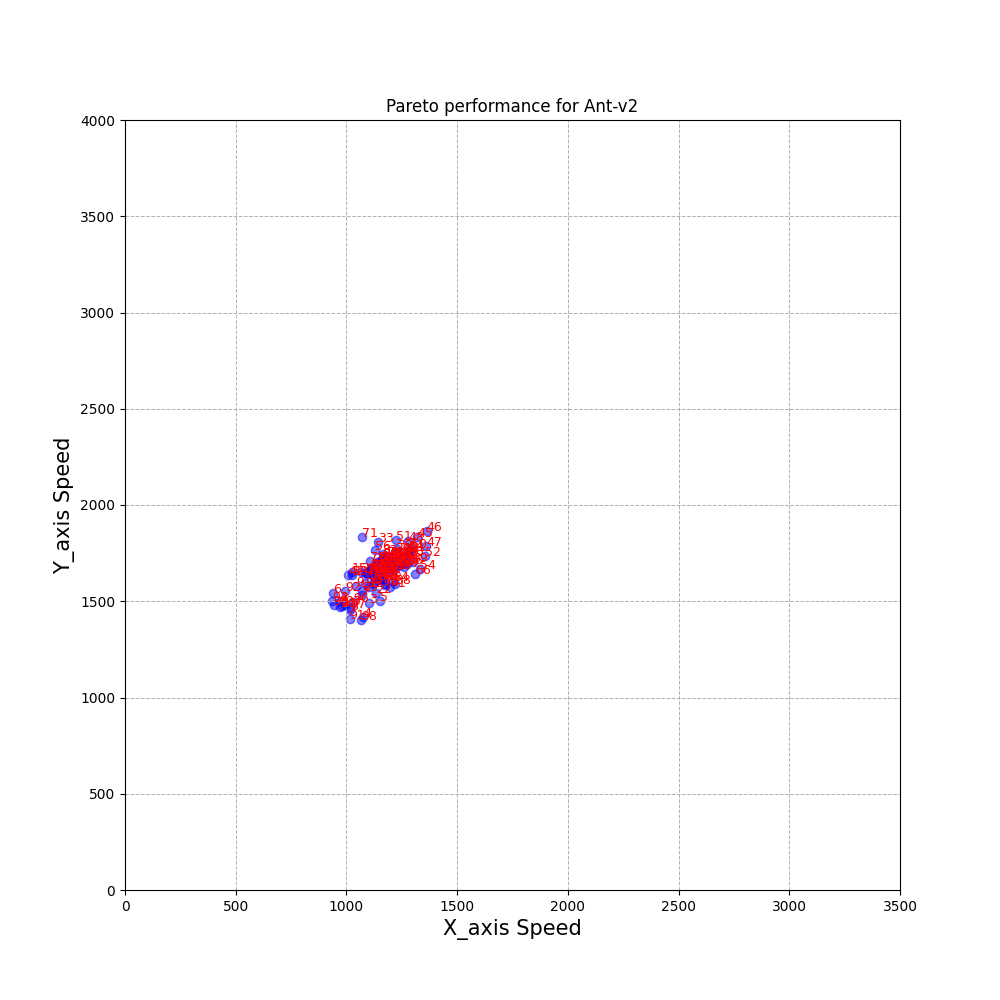}
  \label{fig:sub2}
\end{subfigure}
\begin{subfigure}{0.24\textwidth}
  \centering
  \includegraphics[width=\linewidth]{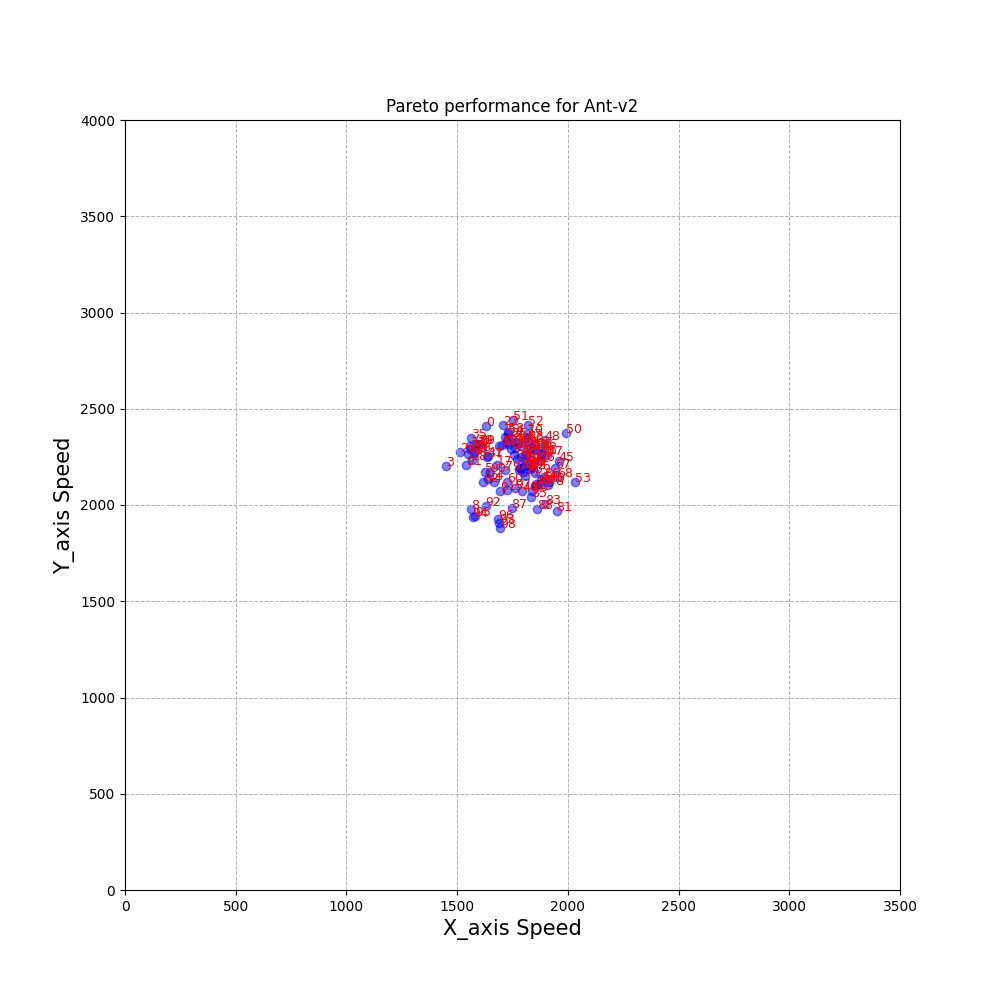}
  \label{fig:sub3}
\end{subfigure}
\begin{subfigure}{0.24\textwidth}
  \centering
  \includegraphics[width=\linewidth]{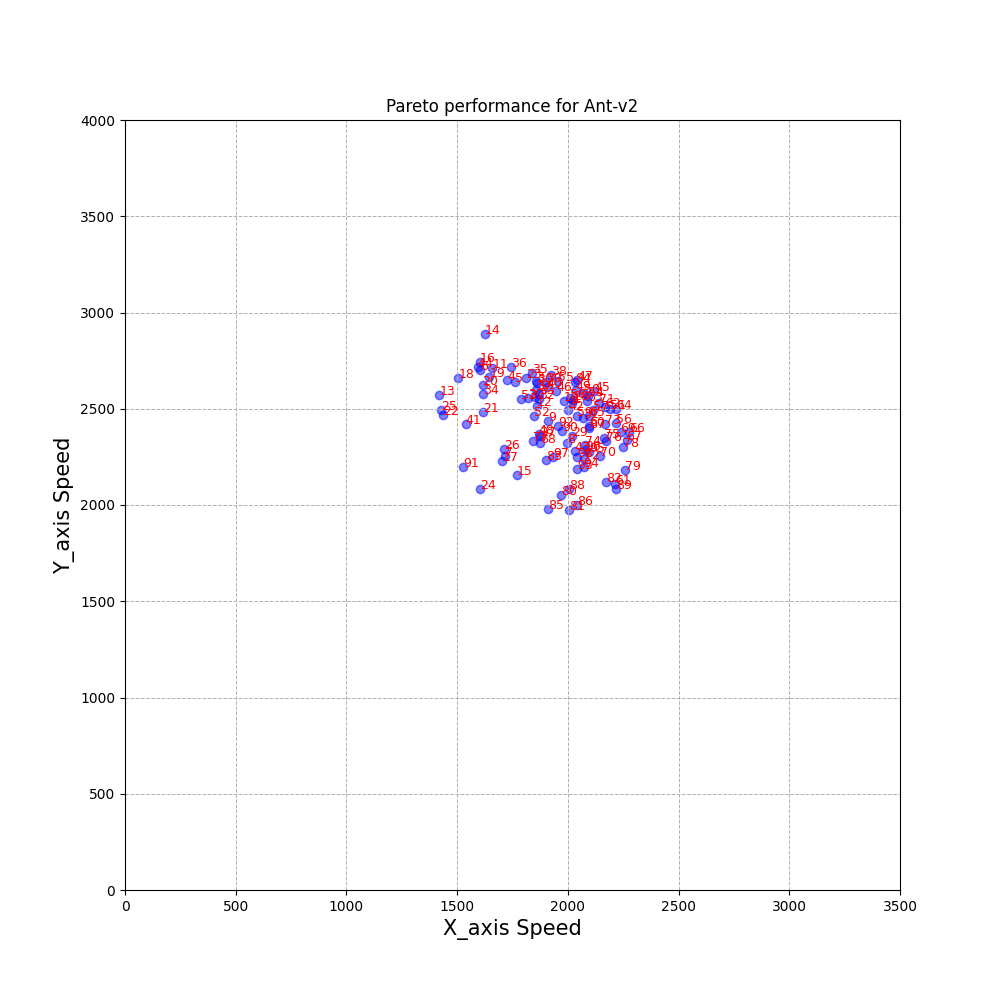}
  \label{fig:sub4}
\end{subfigure}

% Second Row
\begin{subfigure}{0.24\textwidth}
  \centering
  \includegraphics[width=\linewidth]{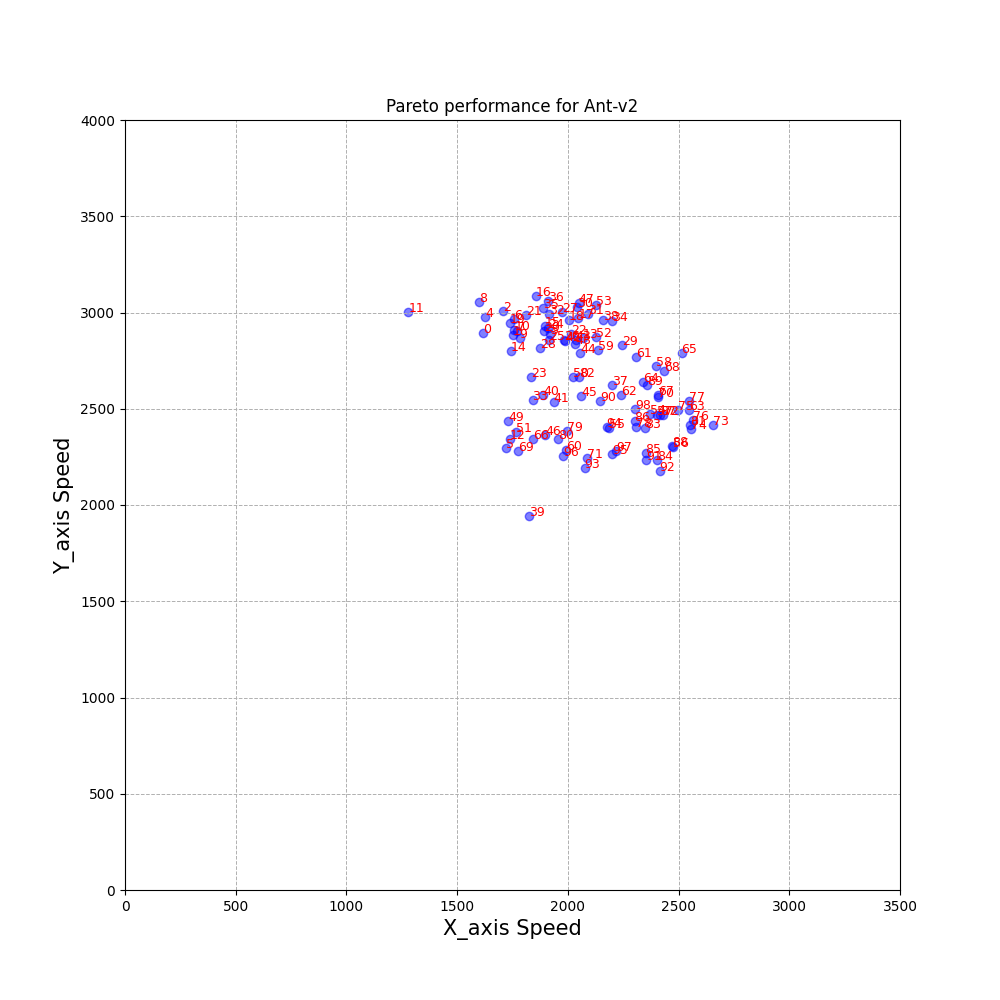}
  \label{fig:sub5}
\end{subfigure}
\begin{subfigure}{0.24\textwidth}
  \centering
  \includegraphics[width=\linewidth]{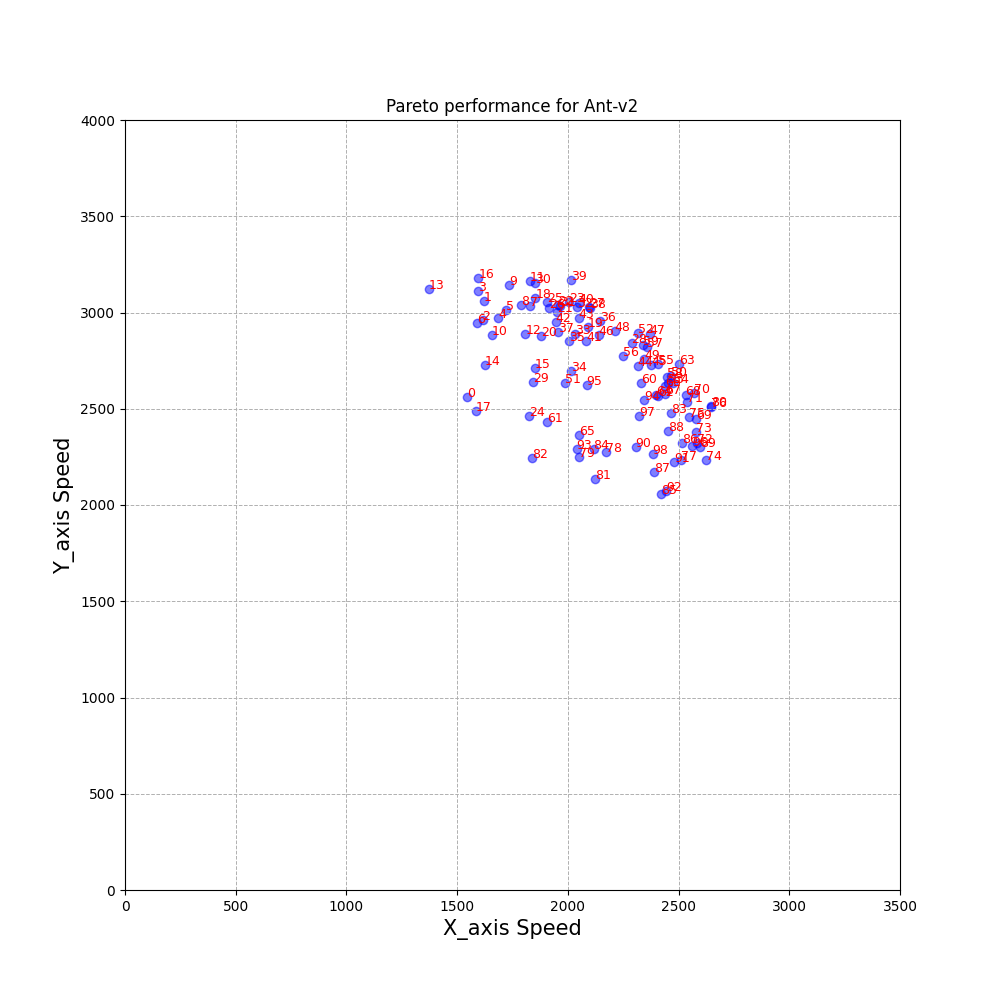}
  \label{fig:sub6}
\end{subfigure}
\begin{subfigure}{0.24\textwidth}
  \centering
  \includegraphics[width=\linewidth]{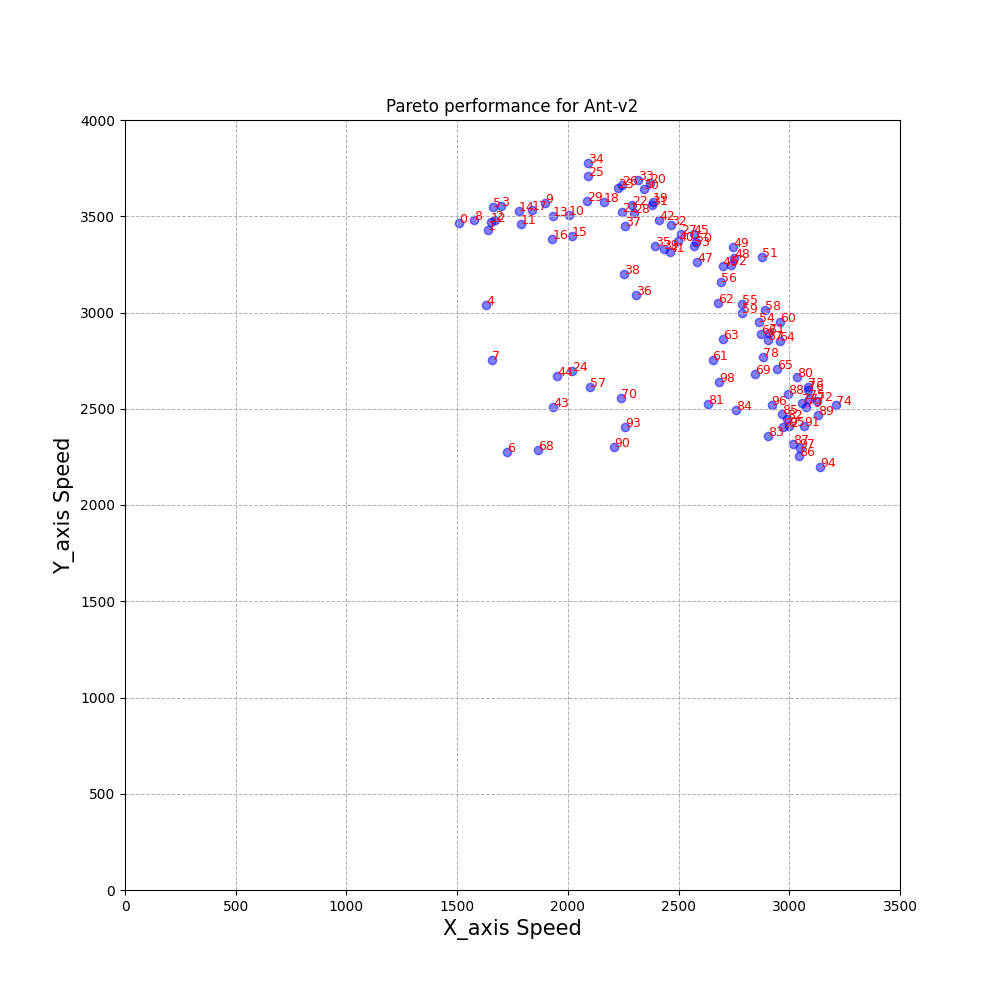}
  \label{fig:sub7}
\end{subfigure}
\begin{subfigure}{0.24\textwidth}
  \centering
  \includegraphics[width=\linewidth]{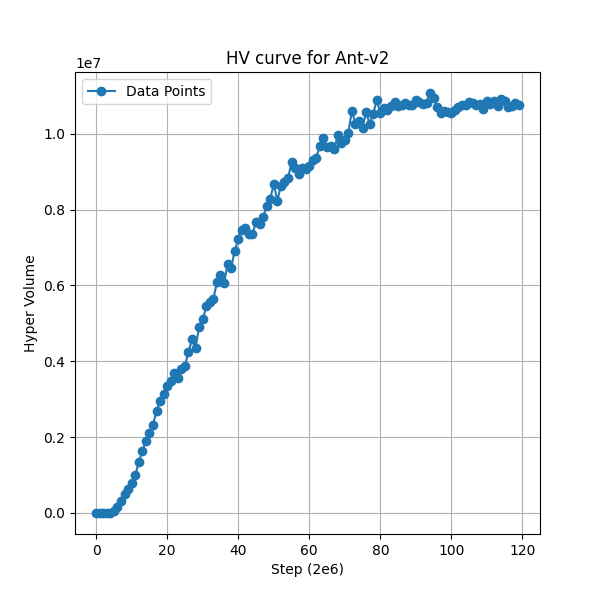}
  \label{fig:sub8}
\end{subfigure}

\caption{CCS expansion in Ant-V2 with one seed. The last graph shows the hypervolume growth.}
\label{fig:ant_ccs}

\end{figure}

% ####################################################################################
% ####################################################################################
\newpage
\subsection{Hopper-V2}
\begin{figure}[H]
\centering
\vspace{-0.5cm}
% First Row
\begin{subfigure}{0.24\textwidth}
  \centering
  \includegraphics[width=\linewidth]{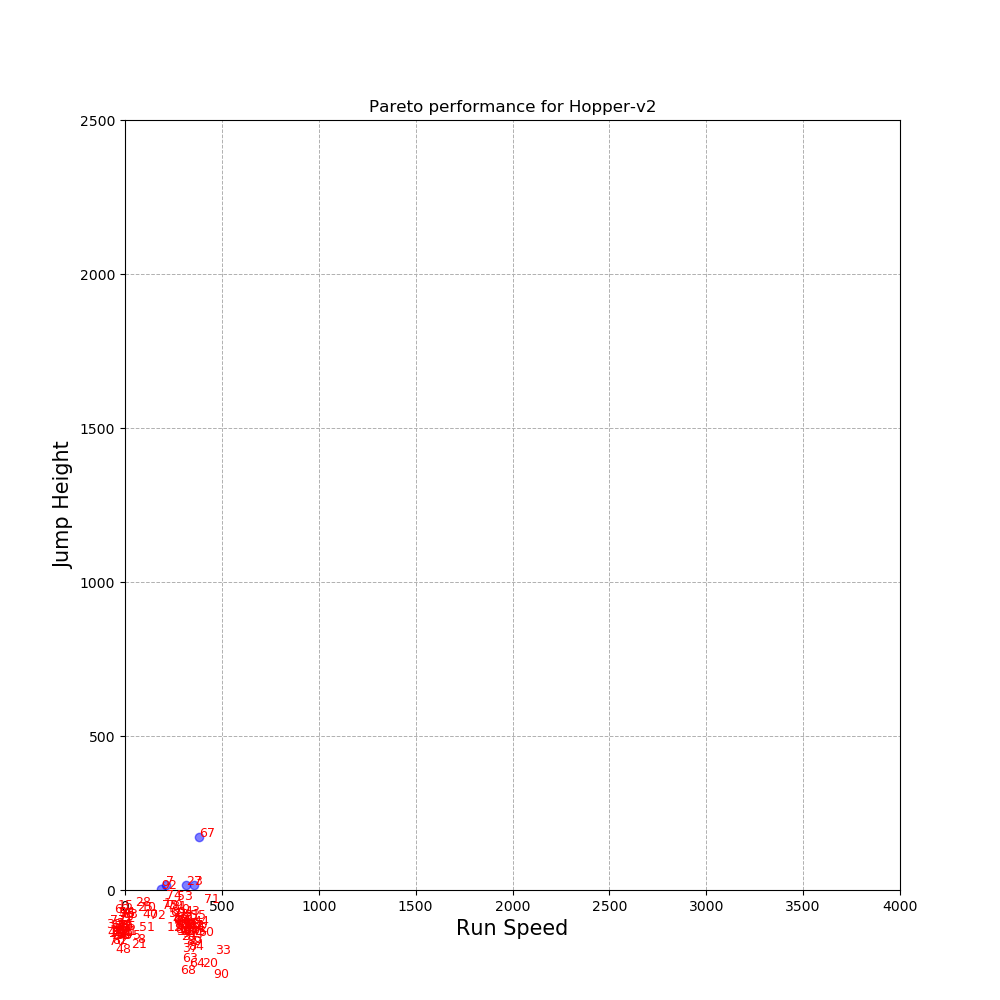}
  \label{fig:sub1}
\end{subfigure}
\begin{subfigure}{0.24\textwidth}
  \centering
  \includegraphics[width=\linewidth]{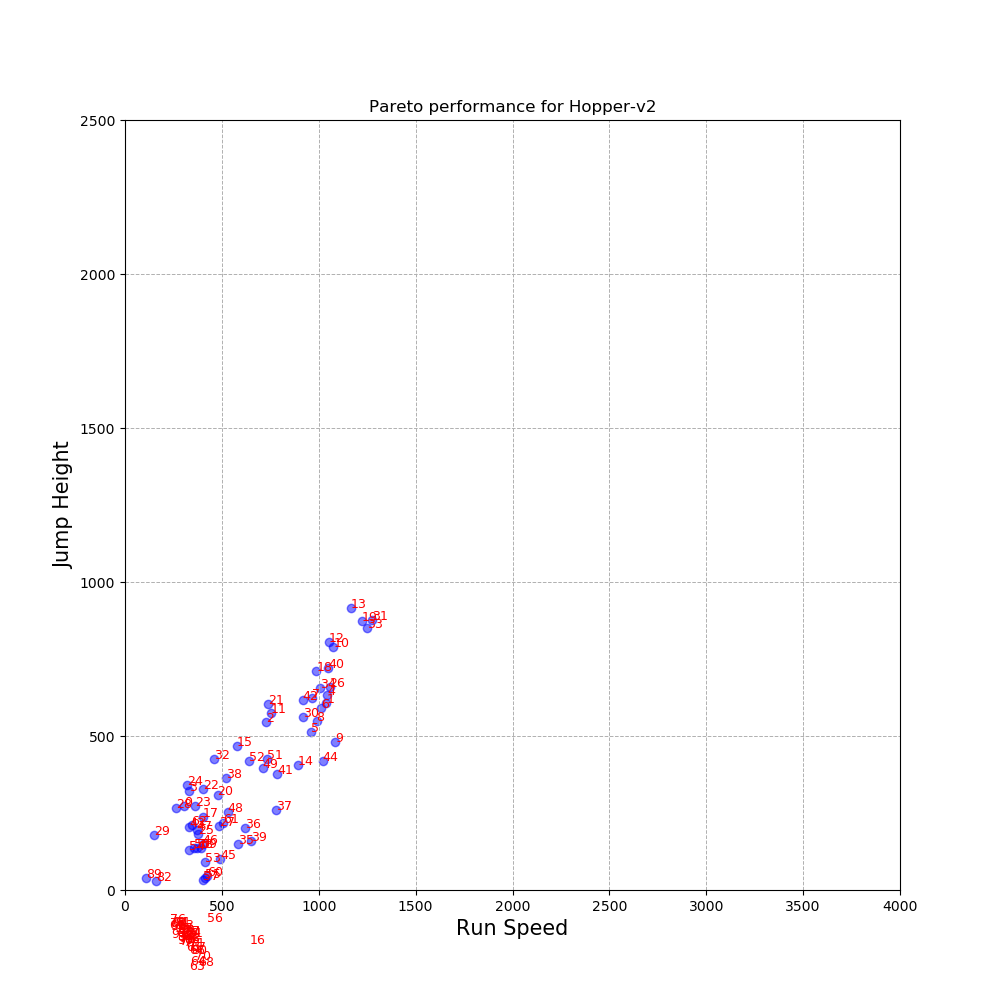}
  \label{fig:sub2}
\end{subfigure}
\begin{subfigure}{0.24\textwidth}
  \centering
  \includegraphics[width=\linewidth]{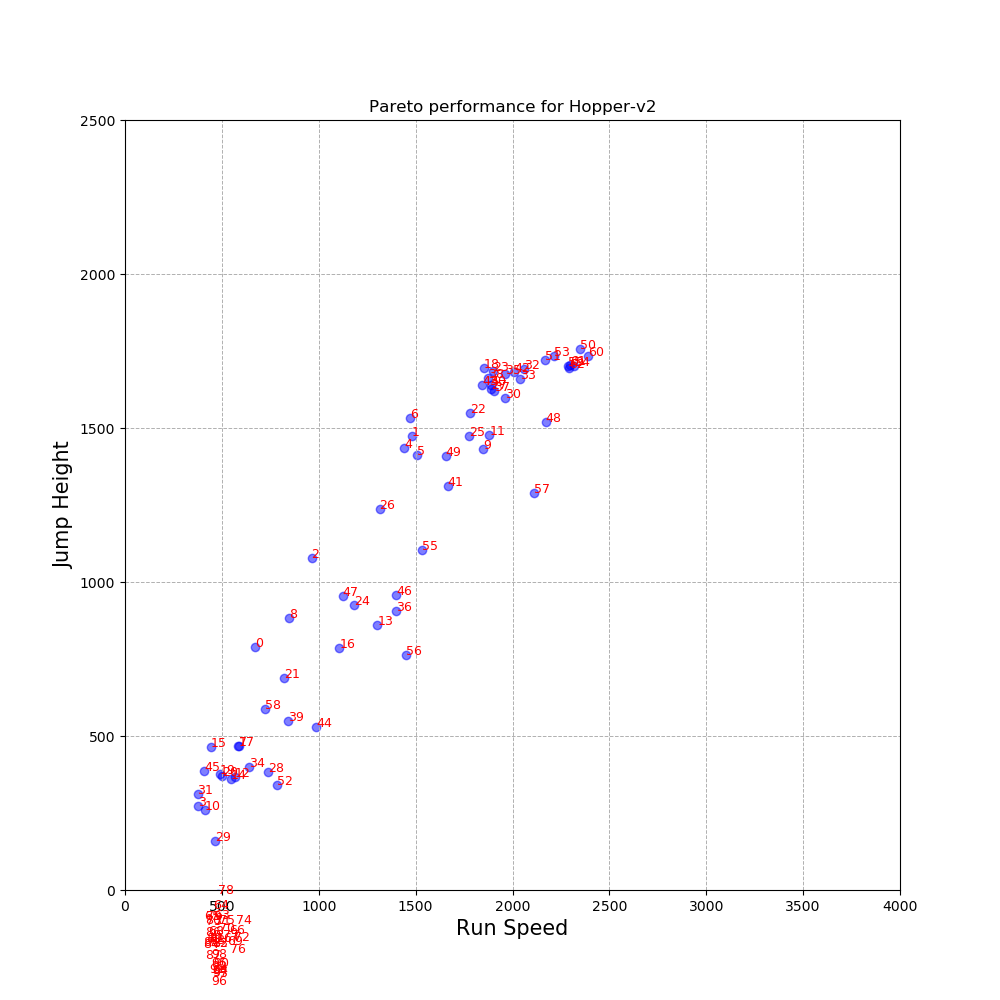}
  \label{fig:sub3}
\end{subfigure}
\begin{subfigure}{0.24\textwidth}
  \centering
  \includegraphics[width=\linewidth]{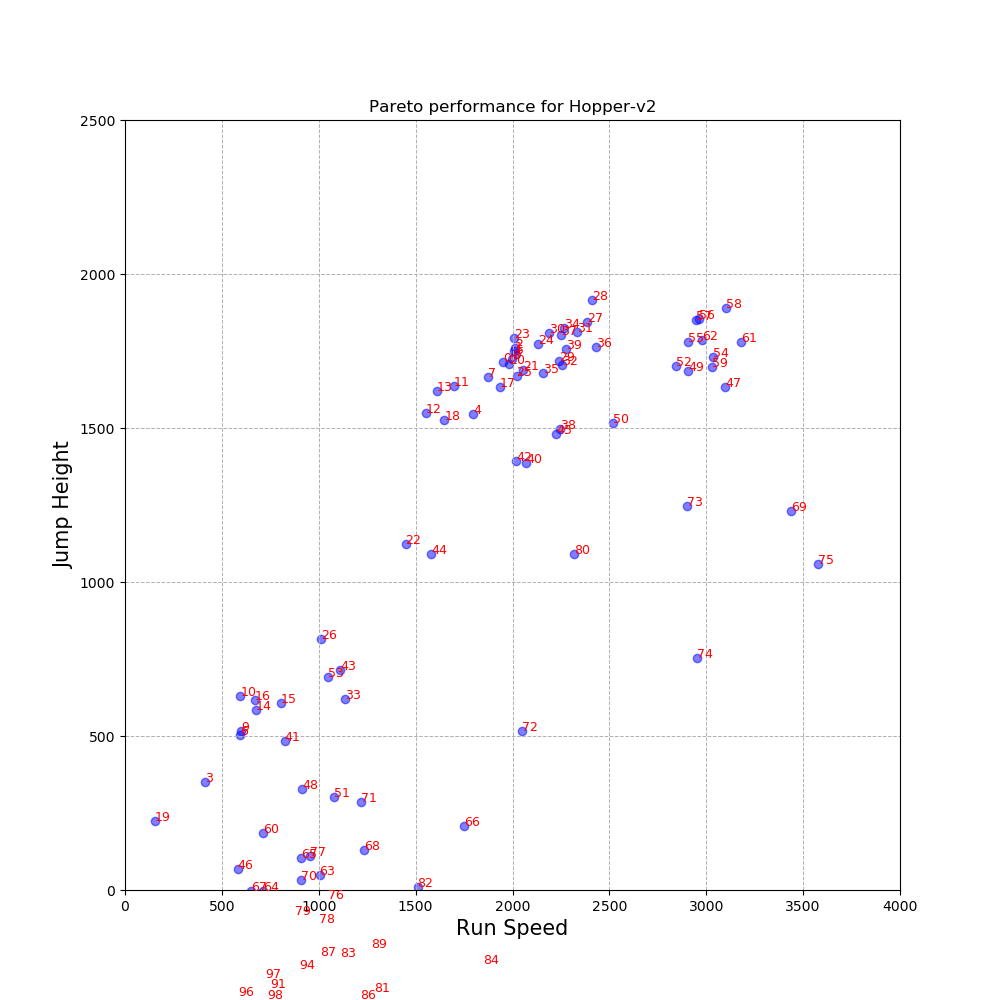}
  \label{fig:sub4}
\end{subfigure}

% Second Row
\begin{subfigure}{0.24\textwidth}
  \centering
  \includegraphics[width=\linewidth]{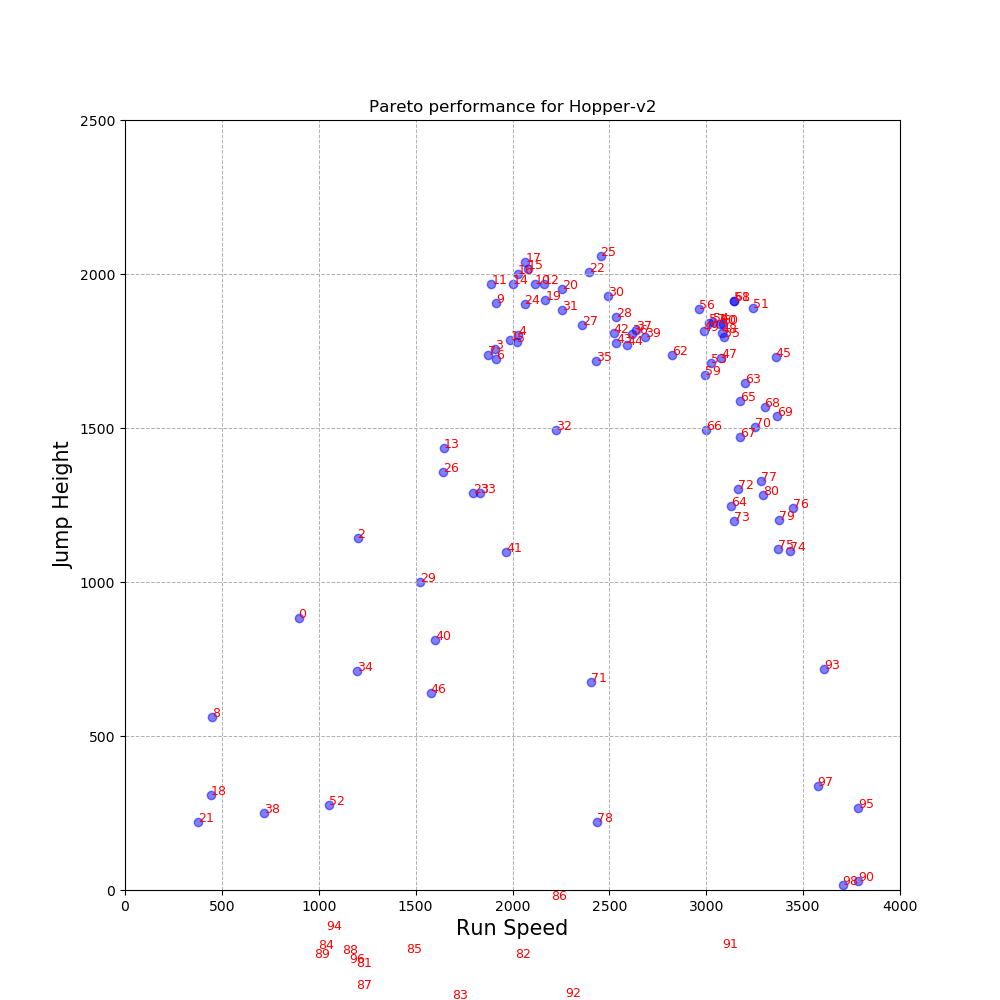}
  \label{fig:sub5}
\end{subfigure}
\begin{subfigure}{0.24\textwidth}
  \centering
  \includegraphics[width=\linewidth]{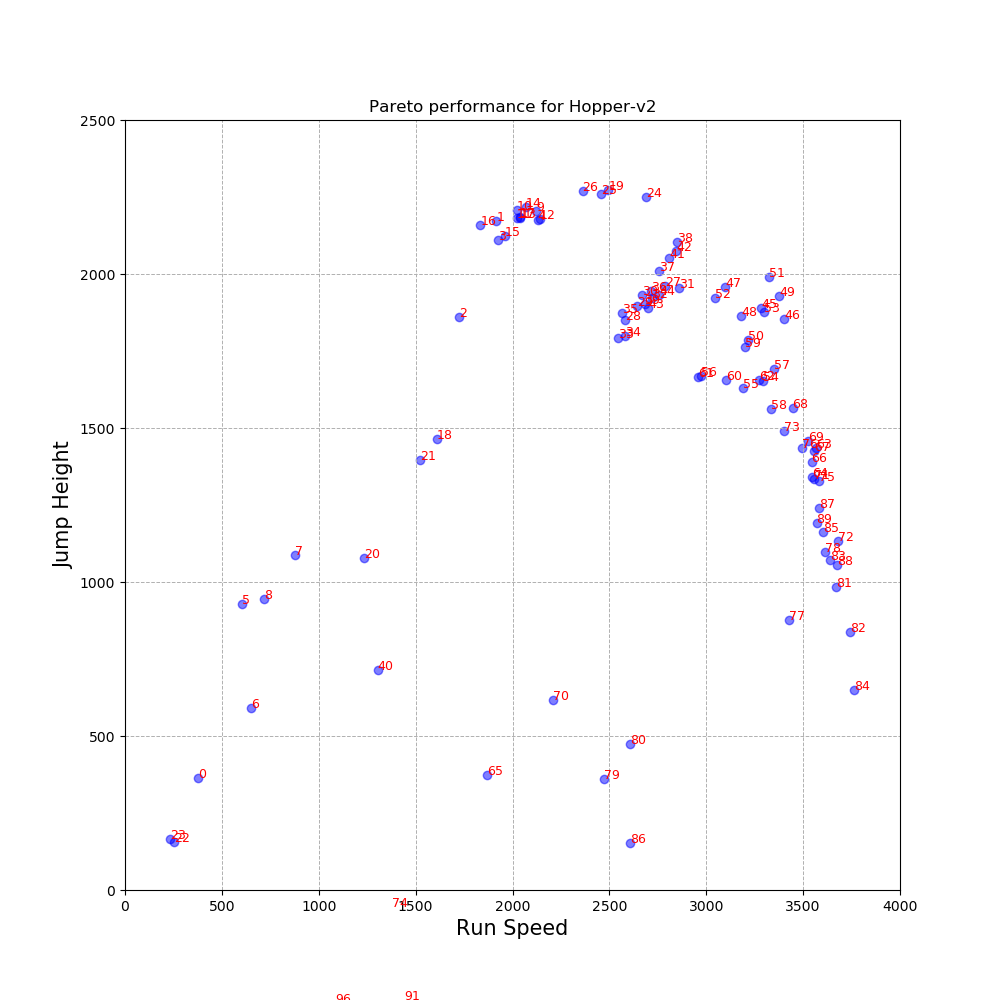}
  \label{fig:sub6}
\end{subfigure}
\begin{subfigure}{0.24\textwidth}
  \centering
  \includegraphics[width=\linewidth]{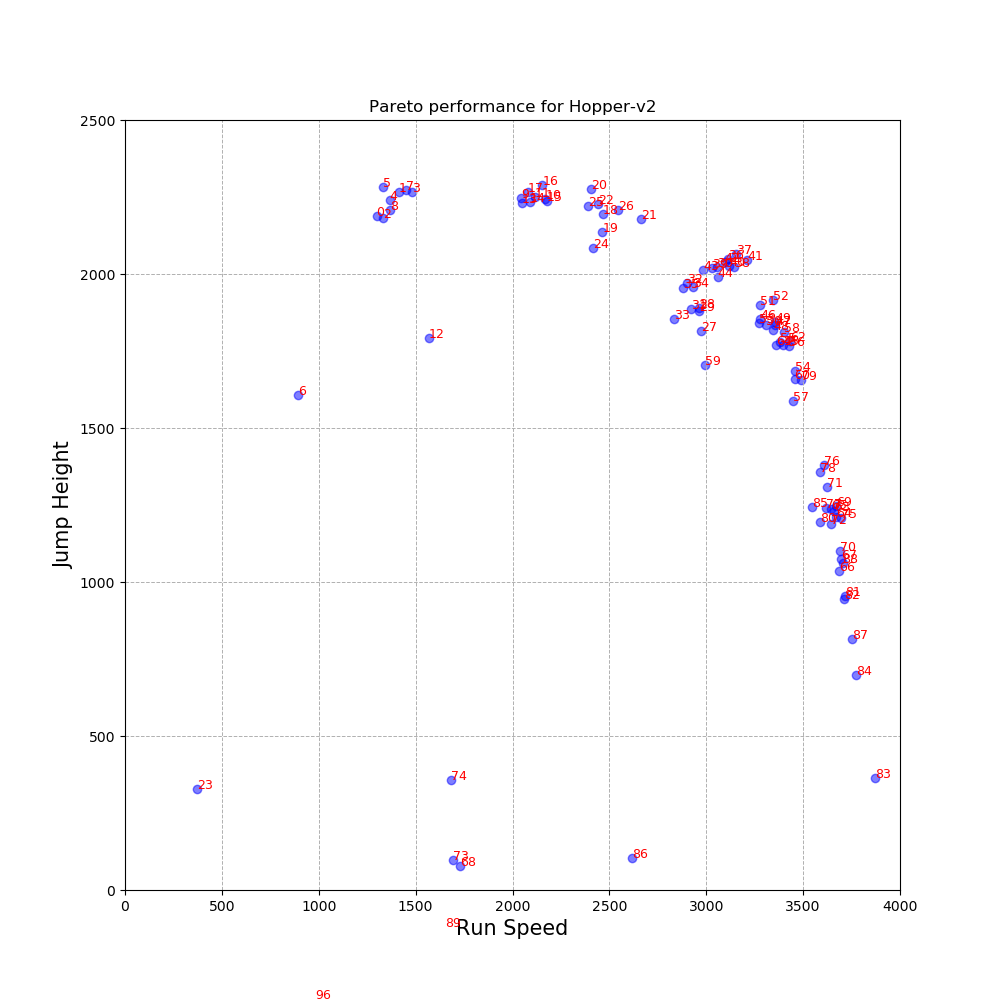}
  \label{fig:sub7}
\end{subfigure}
\begin{subfigure}{0.24\textwidth}
  \centering
  \includegraphics[width=\linewidth]{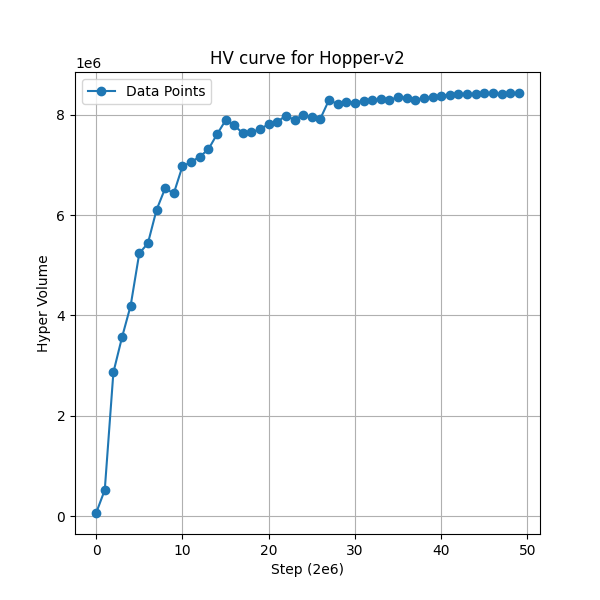}
  \label{fig:sub8}
\end{subfigure}

\caption{CCS expansion in Hopper-V2 with one seed. The last graph shows the hypervolume growth.}
\label{fig:hopper_ccs}

\end{figure}

% ####################################################################################
% ####################################################################################
\newpage
\subsection{Hopper-V3}
\begin{figure}[H]
\centering
\vspace{-0.5cm}
% First Row
\begin{subfigure}{0.24\textwidth}
  \centering
  \includegraphics[width=\linewidth]{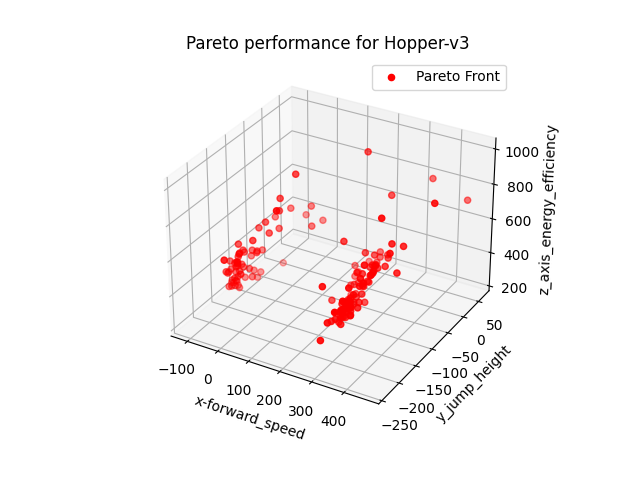}
  \label{fig:sub1}
\end{subfigure}
\begin{subfigure}{0.24\textwidth}
  \centering
  \includegraphics[width=\linewidth]{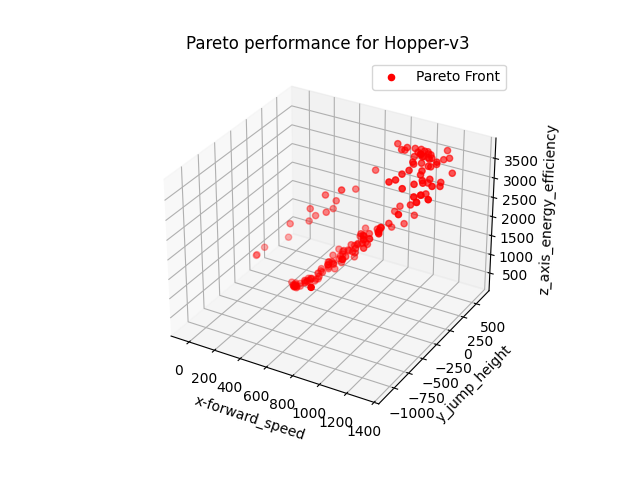}
  \label{fig:sub2}
\end{subfigure}
\begin{subfigure}{0.24\textwidth}
  \centering
  \includegraphics[width=\linewidth]{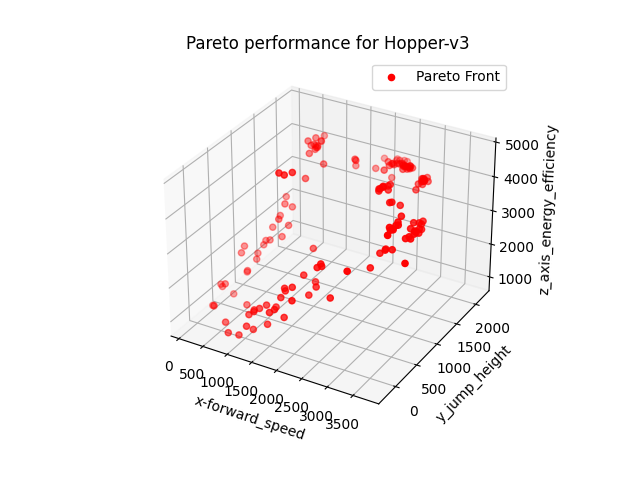}
  \label{fig:sub3}
\end{subfigure}
\begin{subfigure}{0.24\textwidth}
  \centering
  \includegraphics[width=\linewidth]{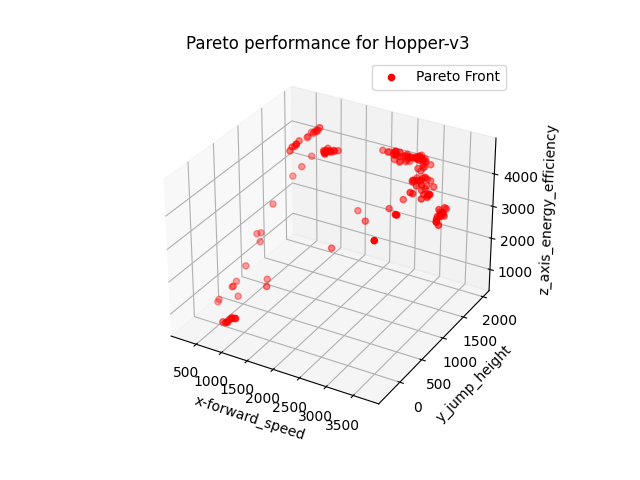}
  \label{fig:sub4}
\end{subfigure}

% Second Row
\begin{subfigure}{0.24\textwidth}
  \centering
  \includegraphics[width=\linewidth]{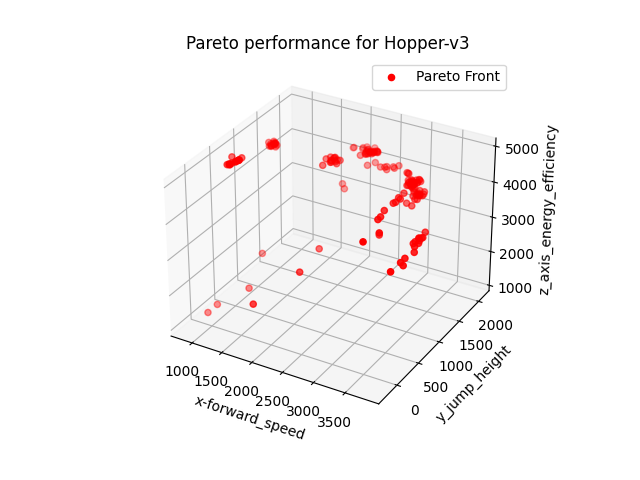}
  \label{fig:sub5}
\end{subfigure}
\begin{subfigure}{0.24\textwidth}
  \centering
  \includegraphics[width=\linewidth]{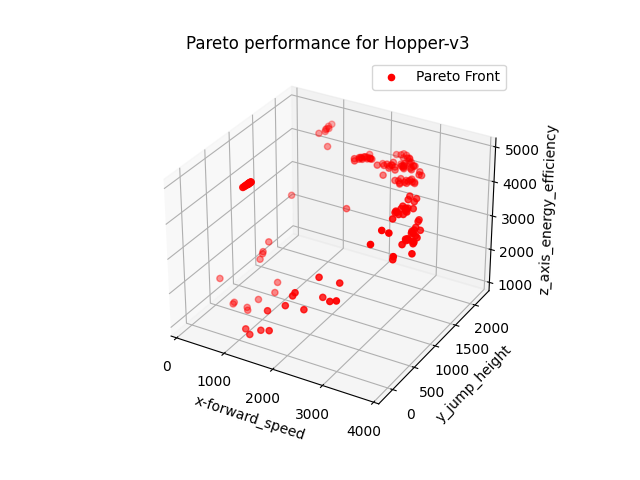}
  \label{fig:sub6}
\end{subfigure}
\begin{subfigure}{0.24\textwidth}
  \centering
  \includegraphics[width=\linewidth]{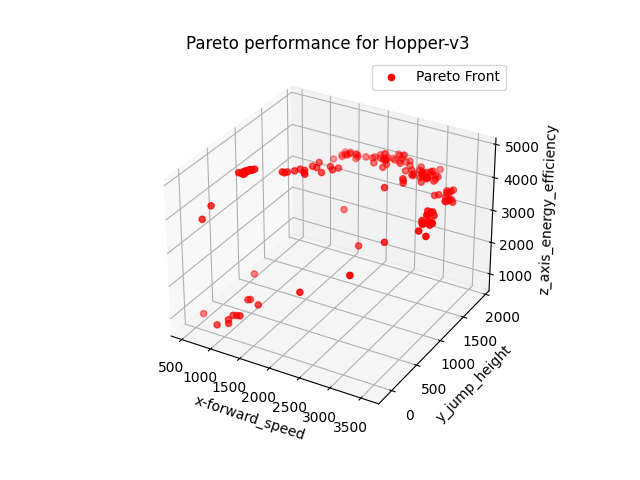}
  \label{fig:sub7}
\end{subfigure}
\begin{subfigure}{0.24\textwidth}
  \centering
  \includegraphics[width=\linewidth]{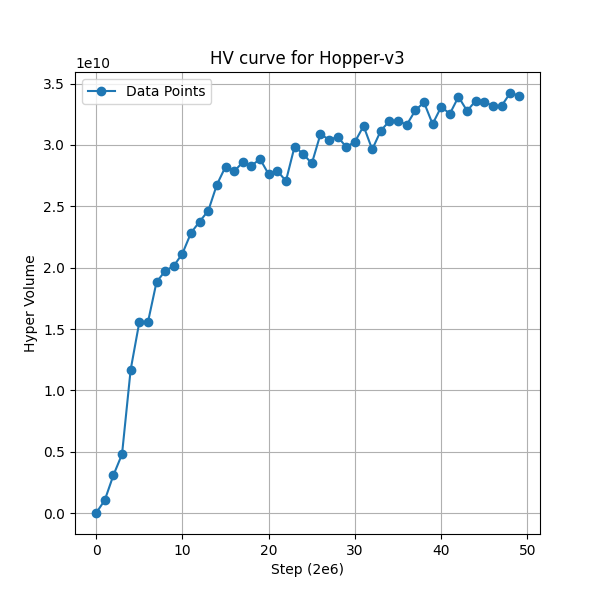}
  \label{fig:sub8}
\end{subfigure}

\caption{CCS expansion in Hopper-V3 with one seed. The last graph shows the hypervolume growth.}
\label{fig:hopperv3_ccs}

\end{figure}

\end{document}